\pdfoutput=1
\documentclass[11pt]{article}

\usepackage[final]{acl}
\usepackage{xurl}
\usepackage{times}
\usepackage{latexsym}
\usepackage{algorithm}
\usepackage{algpseudocode}
\usepackage{tcolorbox}
\usepackage{multirow}
\usepackage{booktabs}
\usepackage{makecell}
\usepackage[T1]{fontenc}
\usepackage[utf8]{inputenc}
\usepackage{amsfonts}

\usepackage{microtype}

\usepackage{inconsolata}

\usepackage{graphicx}

\usepackage[table]{xcolor}
\usepackage{enumitem}

\usepackage{amsmath}
\usepackage{multicol}
\usepackage[normalem]{ulem}
\usepackage{fontawesome5}

\title{Don't Adapt Small Language Models for Tools;\\ Adapt Tool Schemas to the Models} 

\author{Jonggeun Lee$^*$, Woojung Song$^*$, Jongwook Han, Haesung Pyun, Yohan Jo$^{\dag}$ \\
Graduate School of Data Science, Seoul National University \\
\texttt{\{jonggeun.lee, opusdeisong, johnhan00, haesung.pyun, yohan.jo\}@snu.ac.kr}}

\begin{document}
\maketitle

\def\thefootnote{\fnsymbol{footnote}}
\footnotetext[1]{Equal contribution.}
\footnotetext[2]{Corresponding author.}
\def\thefootnote{\arabic{footnote}}

\begin{abstract}

Small language models (SLMs) enable scalable tool-augmented multi-agent systems where multiple SLMs handle subtasks orchestrated by a powerful coordinator.
However, they struggle with tool-use tasks, particularly in selecting appropriate tools and identifying correct parameters. 
A common failure mode is \textit{schema misalignment}: models hallucinate plausible tool names that are absent from the provided tool schema, due to different naming conventions internalized during pretraining. Rather than training models to adapt to unfamiliar schemas, we propose adapting schemas to align with models' pretrained knowledge. 
We introduce \textbf{PA-Tool} (Pretraining-Aligned Tool Schema Generation), a training-free method that leverages peakedness, a signal used in contamination detection that indicates pretraining familiarity, to rename tool components. By generating multiple candidates and selecting the candidate with the highest peakedness, PA-Tool identifies pretraining-aligned naming patterns. Experiments on MetaTool and RoTBench show improvements of up to 17\%, with schema misalignment errors reduced by 80\%.
PA-Tool enables small models to substantially improve tool-use accuracy without retraining, showing that schema-level interventions can unlock the tool-use potential of resource-efficient models. Our code is available at \url{https://github.com/holi-lab/PA-Tool}.
\end{abstract}

\begin{figure}[!t]
    \centering
    \includegraphics[width=0.98\linewidth]{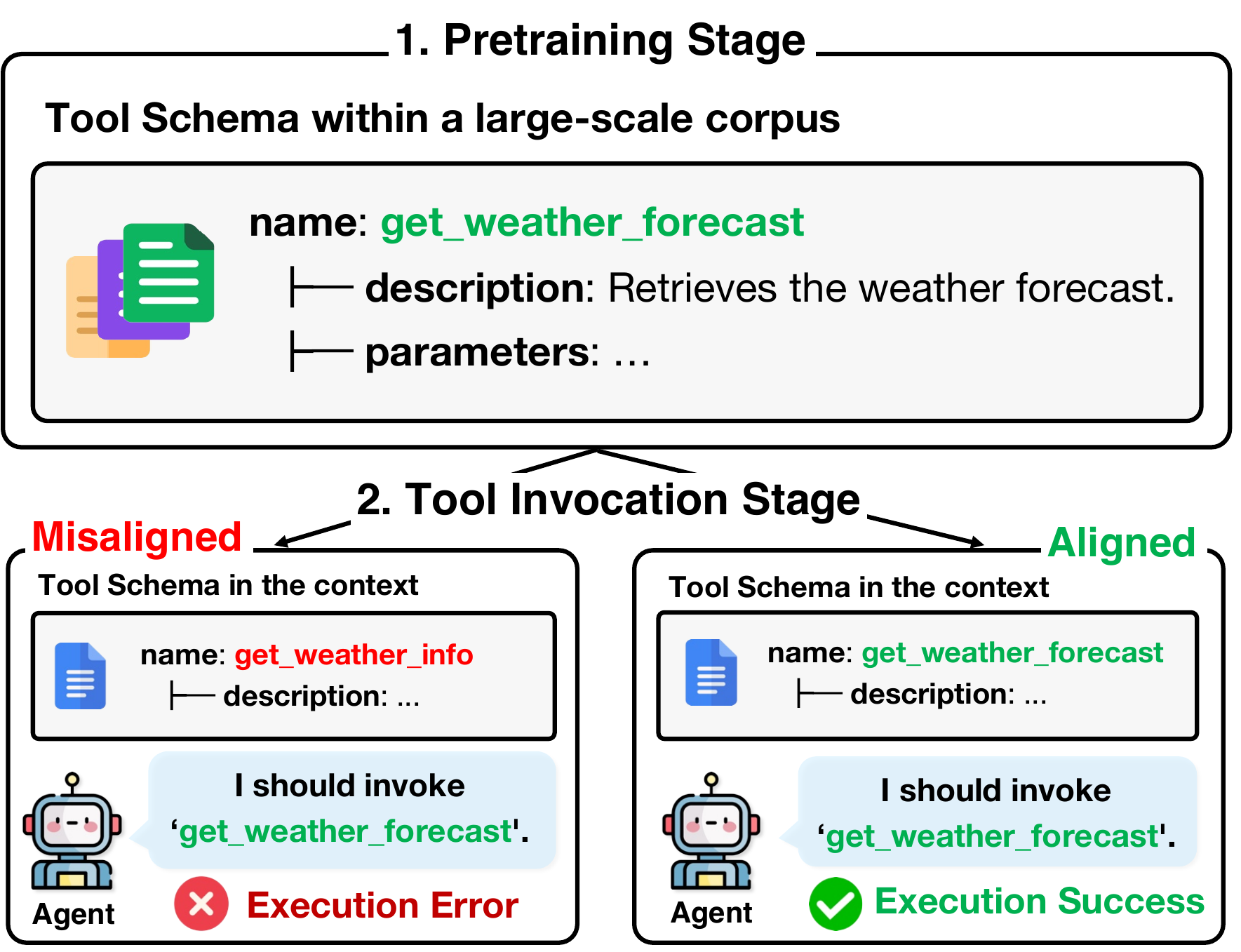}
    \caption{Effect of schema alignment on tool invocation. \textbf{Top}: Models learn tool schemas during pretraining. \textbf{Bottom-Left}: When schemas misalign, models generate plausible but non-existent tools. \textbf{Bottom-Right}: Schema alignment prevents such errors.}
    \label{fig:intro}
    \vspace{-1em}
\end{figure}

%=============================================
\section{Introduction\label{sec:introduction}}
Tool-augmented language models have become essential components of modern AI systems \cite{qu2025toolsurvey}. 
As these systems mature, there is growing interest in deploying small language models (SLMs, typically $\leq 8$B) for tool use. This interest is driven by multi-agent architectures, where a powerful coordinator orchestrates reasoning while multiple SLMs handle subtasks~\cite{belcak2025smalllanguagemodelsfuture}, and by edge deployment scenarios with strict resource constraints~\cite{chen-etal-2025-octopus}.
In these systems, SLMs must perform two critical operations: tool selection (identifying which API to call) and parameter identification (providing appropriate arguments). However, SLMs struggle significantly with these tasks, exhibiting severe performance degradation compared to larger models \cite{erdogan-etal-2024-tinyagent, patil2025the}. 

A common failure pattern is \textit{schema misalignment}: even when appropriate tools are provided in context, models fail to invoke them and instead hallucinate  plausible-looking components that are not in the provided schema\footnote{In this work, a tool schema refers to a hierarchical structure of tools and parameters with descriptions (Figure~\ref{fig:intro}, Top).} (Figure~\ref{fig:intro}, Bottom-Left). This suggests that models fall back on familiar naming conventions internalized during pretraining when faced with unfamiliar schemas. Therefore, we hypothesize that aligning tool component names with patterns from the model's internalized knowledge can reduce this error pattern (Figure~\ref{fig:intro}, Bottom-Right). 

To operationalize this idea, we propose PA-Tool (Pretraining-Aligned Tool Schema Generation), which generates a mapping between original tool component names and alternatives familiar to the models. To identify what naming conventions the model is familiar with, we leverage an insight from contamination detection method \cite{dong-etal-2024-generalization}, which detect whether models encountered specific data during training. A key finding from this study is that patterns frequently seen during training exhibit \textit{peakedness}: models generate highly similar outputs across multiple samples, creating a concentrated distribution. We adopt this \textit{peakedness} as a signal for the model's familiarity with tool names.
PA-Tool renames each component through three stages: (1) instructing the tool-using language model to generate multiple candidate names based on a description of the component \cite{DBLP:journals/corr/abs-2107-03374}, (2) computing the peakedness of each candidate, measured as the number of other candidates sufficiently similar under character-level edit distance \cite{Levenshtein1965BinaryCC}, and (3) selecting the candidate with the highest peakedness as the new name, as it is considered to best align with the model's internalized knowledge.

Experiments on MetaTool and RoTBench demonstrate PA-Tool's effectiveness across diverse scenarios. On MetaTool, PA-Tool achieves tool selection gains of up to 17\% on Reliability (where models must identify that no suitable tool exists) and 9.6\% on multi-tool selection scenarios. On RoTBench, PA-Tool yields tool selection gains of up to 10\% (single-turn) and 6\% (multi-turn), showing that alignment benefits persist across extended contexts, with consistent gains also appearing on parameter identification.

Our comprehensive analysis reveals that PA-Tool's effectiveness stems from directly addressing schema misalignment, the predominant failure mode in SLMs. Errors where models generate plausible but non-existent tool names decrease by 80.0\% with PA-Tool, with accompanying reductions in
other error types (18.8--24.0\%). The underlying mechanism is further validated by showing that peakedness consistently increases as models encounter tool schemas during training, consistent with its role as a familiarity signal.

PA-Tool also demonstrates broad compatibility with existing approaches. For supervised fine-tuning, PA-Tool alone outperforms fine-tuned models on several subtasks without any training, and applying PA-Tool on top of fine-tuned models often yields further improvements. It also provides complementary gains when combined with training-free methods such as retrieval-based correction, constrained generation, and description enhancement. Beyond tool selection benchmarks, PA-Tool improves end-to-end task completion on API-Bank and $\tau$-Bench, confirming that schema alignment translates to practical performance gains.

This training-free approach requires only a one-time schema mapping: rather than forcing models to conform to arbitrary schemas, we adapt the interface to match their internalized knowledge, enabling diverse tool use without model modification, retraining, or risk of catastrophic forgetting.

Our contributions are summarized as follows:
\begin{itemize}
    \item We propose PA-Tool, a training-free schema optimization method that identifies pretraining-aligned tool names by repurposing peakedness, a signal from data contamination.
    \item We demonstrate improvements of up to 17\% across SLMs on MetaTool and RoTBench, with benefits extending from tool selection to parameter identification in both single-turn and multi-turn settings.
    \item We demonstrate that through a simple, easily deployable mapping interface, SLMs can boost their tool-use accuracy while maintaining computational efficiency in resource-constrained systems.
\end{itemize}

%=============================================
\section{Related Work\label{sec:related_works}}

\subsection{Emerging Agentic Frameworks}
Recent advances in LLMs have enabled sophisticated agentic frameworks that decompose complex tasks into specialized modules, each coordinated by dedicated agents \cite{shinn2023reflexion, agashe2025agent, Sapkota_2026}. In these multi-agent systems, SLMs are increasingly replacing larger models within individual modules to reduce computational costs and latency while maintaining specialized functionalities \cite{cheng-etal-2024-small, belcak2025smalllanguagemodelsfuture}. However, these systems remain vulnerable to cascading failures when SLM agents malfunction in foundational tool interaction tasks such as tool selection and parameter identification \cite{erdogan-etal-2024-tinyagent, patil2025the}. A particularly challenging failure mode is schema misalignment: SLM errors often manifest as plausible yet incorrect outputs, such as generating tool names that seem reasonable but do not exist in the actual schema. However, existing works primarily focus on improving agent architectures or coordination strategies, without addressing the schema-level mismatch that causes these failures.

\subsection{Improving Tool Utilization in LLMs} % 
As LLMs demonstrate the capability to interact with external tools \cite{schick2023toolformer,hsieh2023tool}, research has focused on evaluating and improving their tool-use capabilities. Evaluation efforts have developed benchmarks ranging from fine-grained assessments of tool selection and parameter identification \cite{huang2024metatool,li-etal-2023-api,ye-etal-2024-rotbench,chen-etal-2024-eval,patil2025the} to end-to-end multi-step evaluation \cite{qin2024toolllm, trivedi-etal-2024-appworld,yao2025taubench,shim2025noncollaborativeusersimulatorstool, seo2025simuhometemporalenvironmentawarebenchmark}.

Approaches to improving tool-use capabilities follow two main directions. Training-based methods employ supervised fine-tuning \cite{qin2024toolllm, liu2025toolace, zhang-etal-2025-xlam} or reinforcement learning \cite{shi-etal-2024-direct, qian2025toolrl, chen2025reinforcementlearninglonghorizoninteractive, feng2026retool}, but they require substantial data or computational resources. Training-free methods refine tool documentation \cite{yuan-etal-2025-easytool,qu2025from} or leverage interaction histories \cite{fu2024autoguide, wang-etal-2024-llms-imaginarium, zhao2024expel, cui-etal-2025-enhancing-tool}. However, existing training-free approaches primarily focus on improving descriptions or accumulating experiential knowledge through interactions rather than aligning the schema itself with model-preferred representations. Our work directly addresses this gap by generating schemas aligned with models' pretrained knowledge, targeting the root cause of schema misalignment.

\subsection{Data Contamination Detection}

Data contamination, defined as overlap between pretraining data and evaluation benchmarks, can inflate performance through memorization. Early detection methods use n-gram overlap \cite{NEURIPS2020_1457c0d6}, but these approaches fail to detect semantic paraphrasing. Probability-based approaches such as Min-k\% Prob \cite{shi2024detecting} and perplexity analysis \cite{li2023estimatingcontaminationperplexityquantifying} require token probabilities that are unavailable in closed-source models. LLM Decontaminator \cite{yang2023rethinkingbenchmarkcontaminationlanguage} uses auxiliary models for semantic similarity but introduces additional dependencies.
Most relevant to our work, CDD \cite{dong-etal-2024-generalization} identifies memorized patterns by measuring peakedness in sampled candidates, making it applicable to any black-box model. We adapt this peakedness mechanism to identify pretraining-aligned schema patterns, transforming contamination detection into a constructive tool for schema optimization.

%=============================================

\begin{figure*}[!t]
    \centering
    \includegraphics[width=1\textwidth]{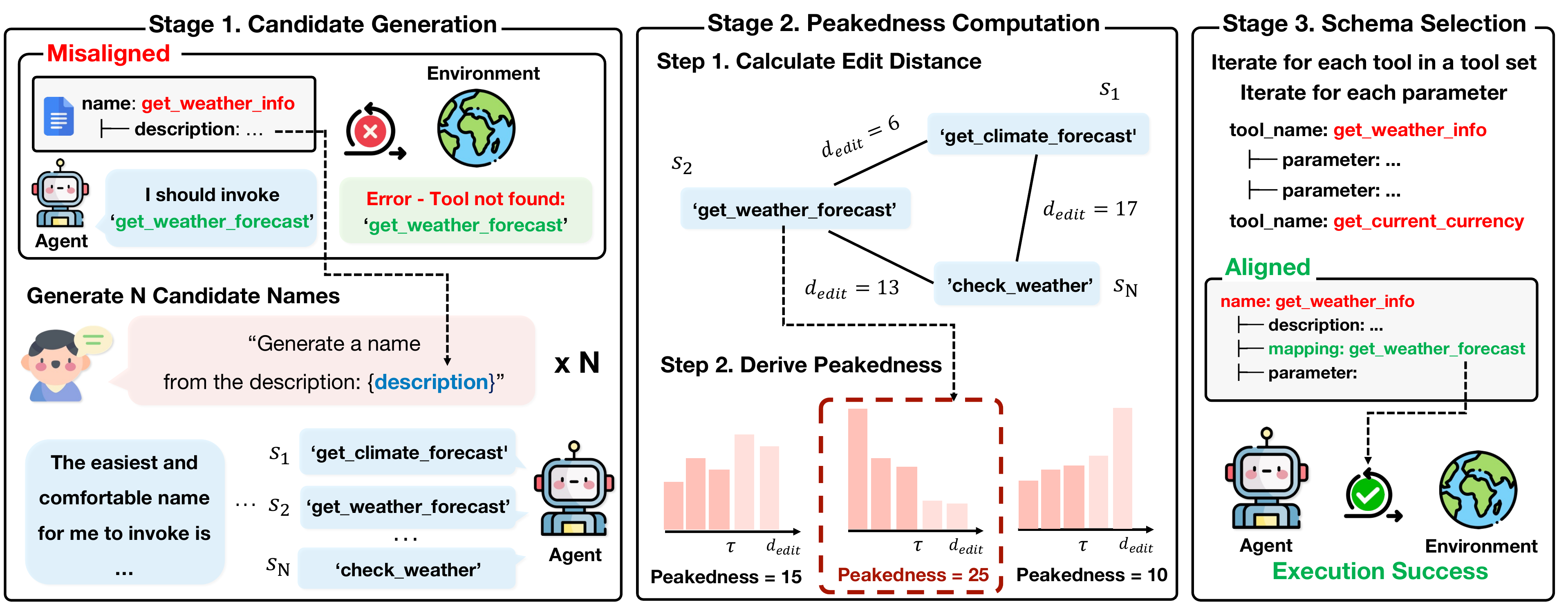}
    \caption{Overview of our PA-Tool framework.}
    
    \label{fig_overview}
    \vspace{-1em}
\end{figure*}
%=============================================
\section{PA-Tool: Pretraining-Aligned Tool Schema Generation}

We now formalize our approach to generating pretraining-aligned tool schemas. A tool schema is structured documentation that defines tools and parameters along with descriptions for each component. Our objective is to rename these components (i.e., tool names and parameter names) with ones that the model has frequently encountered during pretraining.

\subsection{Framework Overview}

Let $\mathcal{M}$ denote a pretrained language model. Given a natural language description $d$ of a schema component's functionality (e.g., a tool or a parameter description), our objective is to generate an optimal name $s^*$ that represents the naming pattern most deeply internalized by $\mathcal{M}$. As illustrated in Figure~\ref{fig_overview}, our approach operates in three stages: (1) we provide $\mathcal{M}$ with a description of a component and instruct it to generate its name multiple times (Stage 1), (2) we compute peakedness scores that measure how many similar candidates cluster around each candidate name (Stage 2), and (3) we select the candidate with the highest peakedness as the pretraining-aligned name (Stage 3). By applying this process to each component in the schema hierarchy, we construct a dictionary mapping from original names to pretraining-aligned names. The complete algorithm is provided in Appendix~\ref{app:patool_algorithm}.

\subsection{Stage 1: Candidate Generation}
In this stage, we collect diverse candidate names of a component that the model may have encountered during pretraining. As illustrated in Figure~\ref{fig_overview}-1, given a component's description, we sample $N$ candidate names $\mathcal{C} = \{s_1, s_2, \ldots, s_N\}$ from the language model with temperature $t \in (0,1]$. This temperature-controlled sampling explores the model's learned distribution beyond the single greedy path, revealing diverse naming patterns~\cite{DBLP:journals/corr/abs-2107-03374}. The detailed prompt structure is provided in Appendix \ref{app:prompt}. We generate a reference name $s_{\text{ref}}$ using greedy decoding ($t=0$) for tie-breaking in the selection stage.

\subsection{Stage 2: Peakedness Computation}

Following contamination detection principles \cite{dong-etal-2024-generalization}, we analyze the local concentration of the output distribution to identify strongly memorized patterns. As illustrated in Figure~\ref{fig_overview}-2, for each candidate name $s_i$, we compute its peakedness score by counting how many similar candidate names the model generates. The intuition is that candidate names with many similar variants indicate patterns frequently encountered during pretraining that the model generates consistently, while isolated candidates suggest less familiar patterns.

To quantify this clustering behavior, we first define a similarity threshold $\tau$ based on the maximum character length in the candidate set:

\begin{equation}
\tau = \alpha \cdot \ell_{\max}
\label{eq:threshold}
\end{equation} where $\ell_{\max} = \max_{s_i \in \mathcal{C}} |s_i|$ is the maximum character length across all candidates, and $\alpha \in [0, 1]$ is a hyperparameter that controls the strictness of similarity. This length-adaptive threshold ensures that longer names are allowed proportionally more variation while maintaining consistent similarity criteria across different name lengths.

The peakedness score for each candidate $s_i$ is then computed as:

\begin{equation}
\phi(s_i) = \sum_{j \neq i} \mathbb{I}(d_{\text{edit}}(s_i, s_j) \leq \tau)
\label{eq:peakedness}
\end{equation} where $d_{\text{edit}}(\cdot, \cdot)$ denotes the character-level edit distance using the Levenshtein algorithm \cite{Levenshtein1965BinaryCC}, and $\mathbb{I}(\cdot)$ is the indicator function that equals 1 when the condition is satisfied and 0 otherwise. This score counts the number of candidates that fall within the similarity threshold from $s_i$, reflecting how many similar names the model generates around this pattern.

\subsection{Stage 3: Schema Selection}

The representative name is selected as the candidate with the maximum peakedness:

\begin{equation}
s^* = \underset{s_i \in \mathcal{C}}{\mathrm{arg\,max}} \; \phi(s_i)
\label{eq:selection}
\end{equation}

This criterion identifies the naming pattern that the model generates most consistently across multiple sampling attempts, suggesting that this pattern is the most deeply internalized from the training data. When multiple candidates tie for maximum peakedness, we select the one with the minimum edit distance to the reference name:

\begin{equation}
s^* = \underset{s_i \in \mathcal{C}^*}{\mathrm{arg\,min}} \; d_{\text{edit}}(s_i, s_{\text{ref}})
\label{eq:tiebreak}
\end{equation} where $\mathcal{C}^* = \{s_i \in \mathcal{C} : \phi(s_i) = \max_{s_j \in \mathcal{C}} \phi(s_j)\}$ contains all candidates with maximum peakedness.

This approach rests on the hypothesis that frequently occurring patterns in training data create local maxima in the model's output distribution. By identifying regions of high peakedness, we effectively locate these memorized naming conventions, which represent the most natural and well-formed component names according to the model's learned knowledge. Through iterative application across all components in a schema (Figure~\ref{fig_overview}-3), we obtain the final pretraining-aligned schema. An example of a schema generated through this process is provided in Appendix~\ref{app:schema_example}. 
We repeat our three-stage process for all tool sets within the system. When name collisions occur across different tools due to highly similar descriptions, we resolve them using the priority-based locking mechanism (Appendix~\ref{app:name_collision}).

%=============================================
\section{Experimental Setup}
\begin{table*}[ht!]
\centering
\small
\resizebox{\textwidth}{!}{
\begin{tabular}{ll|cccc|cccc}
\toprule
\multirow{3}{*}{\textbf{Model}} & \multirow{3}{*}{\textbf{Method}} & \multicolumn{4}{c|}{\textbf{MetaTool}} & \multicolumn{4}{c}{\textbf{RoTBench}} \\
\cmidrule(lr){3-6} \cmidrule(lr){7-10}
& & \multicolumn{4}{c|}{Tool Selection} & \multicolumn{2}{c}{Single-turn} & \multicolumn{2}{c}{Multi-turn} \\
\cmidrule(lr){3-6} \cmidrule(lr){7-8} \cmidrule(lr){9-10}
& & Similar & Scenario & Reliability & Multi-tool & Tool Sel. & Param Iden. & Tool Sel. & Param Iden. \\
\midrule
\rowcolor{gray!15}
\multicolumn{10}{c}{\textit{Closed-Source Models}} \\
\midrule
GPT-4.1-mini & Base & \textbf{79.6} & \textbf{84.3} & 76.3 & 72.2 & 79.1 & 58.1 & \textbf{71.4} & \textbf{61.4} \\
Gemini-2.5-Flash & Base & 70.0 & 79.8 & \textbf{89.2} & 77.3 & 82.9 & 56.2 & 60.0 & 54.3 \\
Claude-Sonnet-4.5 & Base & 75.6 & 83.0 & 84.3 & \textbf{85.1} & \textbf{83.3} & \textbf{67.6} & 64.3 & 58.6 \\
\midrule
\rowcolor{gray!15}
\multicolumn{10}{c}{\textit{Small Language Models}} \\
\midrule
\multirow{5}{*}{Qwen2.5-3B}
& Base & 48.7 & 55.3 & 83.6 & \textbf{75.1} & 12.4 & 7.6 & 10.0 & 10.0 \\
& Greedy & 49.1 & 57.0 & 82.9 & 63.4 & 13.3 & 7.6 & 11.4 & 7.1 \\
& MostFreq & 50.5 & 56.3& 84.3& 73.0 & 14.3 & 8.6 & 12.9 & 8.6 \\
& Human & \textbf{54.6}& \textbf{62.9} & 80.6& 72.6& - & - & - & - \\
& PA-Tool & 50.0 & 58.8 & \textbf{86.2} & 72.6 & \textbf{18.1} & \textbf{10.5} & \textbf{15.7} & \textbf{14.3} \\
\midrule
\multirow{5}{*}{Qwen2.5-7B} 
& Base & 59.6 & 74.4 & 78.3 & 78.3 & 49.5 & 20.0 & 21.4 & 21.4 \\
& Greedy & 60.6 & 75.4 & 85.4 & 82.7 & 50.5 & 21.0 & 18.6 & 15.7 \\
& MostFreq & 65.3 & 75.5 & \textbf{90.0} & \textbf{86.1} & 50.5 & \textbf{23.8} & 24.3 & 20.0 \\
& Human & \textbf{65.8} & 76.7 & 78.6 & 82.1 & - & - & - & - \\
& PA-Tool & 64.1 & \textbf{78.4} & 88.2 & 84.9 & \textbf{55.2} & 21.9 & \textbf{27.1} & \textbf{22.9} \\
\midrule
\multirow{5}{*}{Llama3.2-3B}
& Base & 55.0 & 58.6 & 43.6 & 79.1 & 56.2 & 20.0 & 32.9 & 27.1 \\
& Greedy & 57.7 & 58.9 & 39.8 & 70.4 & 59.1 & 20.0 & 32.9 & 25.7 \\
& MostFreq & 61.3 & 64.2 & \textbf{65.7} & \textbf{81.3} & 60.0 & 18.1 & 32.9 & 27.1 \\
& Human & 58.6 & 60.9 & 55.2 & 72.6 & - & - & - & - \\
& PA-Tool & \textbf{65.7} & \textbf{67.7} & 60.6 & 80.5 & \textbf{62.9} & \textbf{21.9} & \textbf{34.3} & \textbf{28.6} \\
\midrule
\multirow{5}{*}{Llama3.1-8B} 
& Base & 61.5 & 73.9 & 53.5 & 78.7 & 58.1 & 17.1 & 42.8 & 34.3 \\
& Greedy & 64.6 & 72.9 & 51.5 & 78.9 & 63.8 & \textbf{18.1} & 44.3 & 32.9 \\
& MostFreq & 68.8 & 79.3 & \textbf{66.4} & 85.7 & 66.7 & \textbf{18.1} & 45.7 & 34.3 \\
& Human & 69.3 & 78.9 & 63.5 & 86.5 & - & - & - & - \\
& PA-Tool & \textbf{70.4} & \textbf{79.9} & 66.0 & \textbf{88.3} & \textbf{68.6} & \textbf{18.1} & \textbf{48.6} & \textbf{35.7} \\
\bottomrule
\end{tabular}
}
\caption{Performance comparison on MetaTool and RoTBench. All metrics are reported as accuracy (\%). \textbf{Bold} indicates the best performance within each open-source model and among all closed-source models.}
\label{tab:model_comparison}
\vspace{-1em}
\end{table*}

\subsection{Benchmarks}

\noindent\textbf{MetaTool.} MetaTool \citep{huang2024metatool} evaluates tool selection capabilities across 4,287 test cases with 199 tools in four subtasks: (1) \textbf{Similar} tests semantic comprehension by distinguishing tools with overlapping functionalities (e.g., Sudoku vs. Tic-Tac-Toe); (2) \textbf{Scenario} selects appropriate tools based on user-specific contexts and requirements (e.g., software engineers, students); (3) \textbf{Reliability} assesses whether models can directly indicate when no suitable tool is available rather than hallucinating; and (4) \textbf{Multi-tool} measures whether models can correctly select multiple tools when tasks require composition of functionalities.

\noindent\textbf{RoTBench.} RoTBench \citep{ye-etal-2024-rotbench} evaluates two tool-use capabilities across 105 test cases with 568 tools: (1) \textbf{Tool Selection} assesses whether models correctly identify the appropriate tool; and (2) \textbf{Parameter Identification} measures whether models accurately extract the required parameter set, conditioned on correct tool selection. We evaluate in both single-turn and multi-turn settings, where the latter provides two preceding interaction turns
(e.g., a failed attempt with incorrect parameters) before evaluating on the third turn. RoTBench also introduces four levels of noise perturbation to tool component names (e.g., reversed names) to stress-test robustness; we focus on the Clean environment in our main experiments, with results across all levels reported in Appendix~\ref{app:noise_levels}.

\subsection{Models and Baselines}
We primarily focus on SLMs, evaluating four open-source models: Qwen2.5-3B/7B \citep{qwen2025qwen25technicalreport}, Llama3.1-8B \citep{meta_llama31_2024}, and Llama3.2-3B \citep{meta_llama32_2024}. For each model, we evaluate five configurations: (1) \textbf{Base} uses the original schema in the benchmarks without modifications; (2) \textbf{Greedy} uses greedy decoding to generate deterministic schema names; (3) \textbf{MostFreq} selects the most frequently generated candidate from our sampling process; (4) \textbf{Human} uses schemas where two PhD-level annotators with software engineering backgrounds manually renamed tools to more intuitive names (evaluated only on MetaTool due to RoTBench's large tool sets); (5) \textbf{PA-Tool} applies our peakedness-based method. To contextualize SLM performance and quantify PA-Tool's improvements, we also compare against three state-of-the-art closed-source models: GPT-4.1-mini \citep{openai_gpt41mini_2024}, Gemini-2.5-Flash \citep{google_gemini25flash_2025}, and Claude-Sonnet-4.5 \citep{anthropic_2025_sonnet}.

\subsection{Implementation Details}
When generating schemas with PA-Tool, we use 32 candidates at temperature 0.4 with $\alpha=0.2$ (sensitivity analysis in Appendix~\ref{app:hyperparameter}). Each component's description is provided individually (see Figure~\ref{fig_overview}-1; prompt template in Appendix~\ref{app:prompt}).

For benchmark inference, we use temperature 0 to ensure reproducible results across all experiments. We use accuracy as the primary evaluation metric across both benchmarks, measuring the percentage of test cases where the model's predictions exactly match the ground-truth labels. Detailed evaluation protocols are in Appendix~\ref{app:evaluation_protocol}.

%=============================================
\section{Main Results}
Table~\ref{tab:model_comparison} presents comprehensive results across MetaTool and RoTBench benchmarks, demonstrating PA-Tool's effectiveness.

\paragraph{MetaTool.}
PA-Tool substantially improves performance over Base models across most MetaTool subtasks. The most substantial gains appear in Reliability (up to 17.0\%, e.g., Llama3.2-3B: 43.6→60.6\%), where models must recognize when no suitable tool exists, a task critically dependent on clearly understanding available options. In Multi-tool, gains reach 9.6\% (Llama3.1-8B: 78.7→88.3\%). When tasks require identifying multiple tools simultaneously, schema misalignment compounds across each selection, making alignment particularly critical. Similar and Scenario tasks show improvements up to 10.7\%.

PA-Tool outperforms training-free alternatives on most tasks. Greedy occasionally underperforms Base (Llama3.2-3B Reliability: 39.8\% vs. 43.6\%) as it generates only a single candidate, limiting schema space exploration. While MostFreq sometimes achieves competitive results by capturing frequency, PA-Tool shows more consistent improvements by measuring distributional concentration. Compared to Human-designed schemas, PA-Tool achieves comparable or superior performance (Llama3.2-3B Similar: 65.7\% vs. 58.6\%), demonstrating automated alignment can match human intuition while being scalable.

\paragraph{RoTBench.} RoTBench evaluates models in single-turn and multi-turn settings across tool selection and parameter identification. PA-Tool demonstrates consistent improvements in single-turn tool selection across all models, with gains ranging from 5.7\% to 10.5\% (Llama3.1-8B: 58.1→68.6\%). In multi-turn settings, improvements of up to 6\% show that alignment benefits persist across extended contexts. Notably, improvements extend beyond tool selection to parameter identification, with gains of up to 4.3\% (Qwen2.5-3B multi-turn: 10.0→14.3\%), confirming that schema alignment improves overall tool-use accuracy.

\paragraph{Comparison with Closed-source Models.}
While a gap remains against closed-source models (Claude-Sonnet-4.5: 83.3\% tool selection on RoTBench single-turn), PA-Tool enables small models to achieve competitive or superior performance on specific subtasks. In MetaTool's Multi-tool, Llama3.1-8B with PA-Tool (88.3\%) surpasses all closed-source models including Claude-Sonnet-4.5 (85.1\%). In Reliability, Qwen2.5-7B with PA-Tool (88.2\%) approaches Gemini-2.5-Flash (89.2\%). These results suggest that targeted schema alignment can narrow the gap with larger models on subtasks where tool naming is a key factor.

%=============================================

\paragraph{Generalization to Diverse Models.}
Beyond the SLMs examined here, we apply PA-Tool to diverse model families and scales, including SLMs (Ministral-8B, GPT-4.1-nano, Gemini-2.5-Flash-Lite), LLMs (Llama3.3-70B, GPT-4.1-mini, Gemini-2.5-Flash), and reasoning models (Qwen3-1.7B and Qwen3-4B in thinking mode). As shown in Table~\ref{tab:additional_models}, PA-Tool improves performance across most models and settings, with larger gains for SLMs where schema misalignment is a dominant failure mode (e.g., Ministral-8B gains 8.5\% on RoTBench single-turn tool selection). For LLMs, gains are smaller on average but remain substantial where misalignment compounds, such as multi-tool composition (e.g., GPT-4.1-mini gains 12.1\% on MetaTool Multi-tool). For reasoning models, PA-Tool provides gains of up to 10.5\% on RoTBench (Qwen3-1.7B single-turn parameter identification) despite strong baselines from chain-of-thought, confirming that reasoning and schema alignment address complementary failure modes (see Appendix~\ref{app:generalization_to_diverse_model} for details).

\section{Analysis\label{sec:analysis}}

We analyze PA-Tool along three directions. We first ask \textit{why} it works, identifying the error modes it addresses, validating that peakedness reflects pretraining familiarity, and isolating which schema components drive the gains (§\ref{sec:error_analysis}--\ref{sec:component_ablation}). We then ask how it relates to existing tool-use approaches, showing that it complements both training-based fine-tuning and other training-free methods rather than competing with them (§\ref{sec:sft}--\ref{sec:integration_with_trainingfree}). Finally, we ask how far its benefits extend, testing generalization to end-to-end benchmarks, human perception, and cross-model schema transfer (§\ref{sec:other_benchmarks}--\ref{sec:cross_model_main}).

\subsection{Error Analysis}\label{sec:error_analysis}

To understand how PA-Tool addresses different failure modes, we analyze error distributions in Llama3.1-8B before and after applying PA-Tool. We categorize incorrect predictions into three types: \textit{Schema Misalignment Error} (generating non-existent but plausible tools), \textit{Functional Confusion Error} (selecting wrong tools with similar functionality), and \textit{Context Understanding Error} (selecting functionally unrelated tools). Detailed definitions and settings are in Appendix~\ref{app:error-analysis}.

\begin{figure}[!t]
    \centering
    \includegraphics[width=0.98\linewidth]{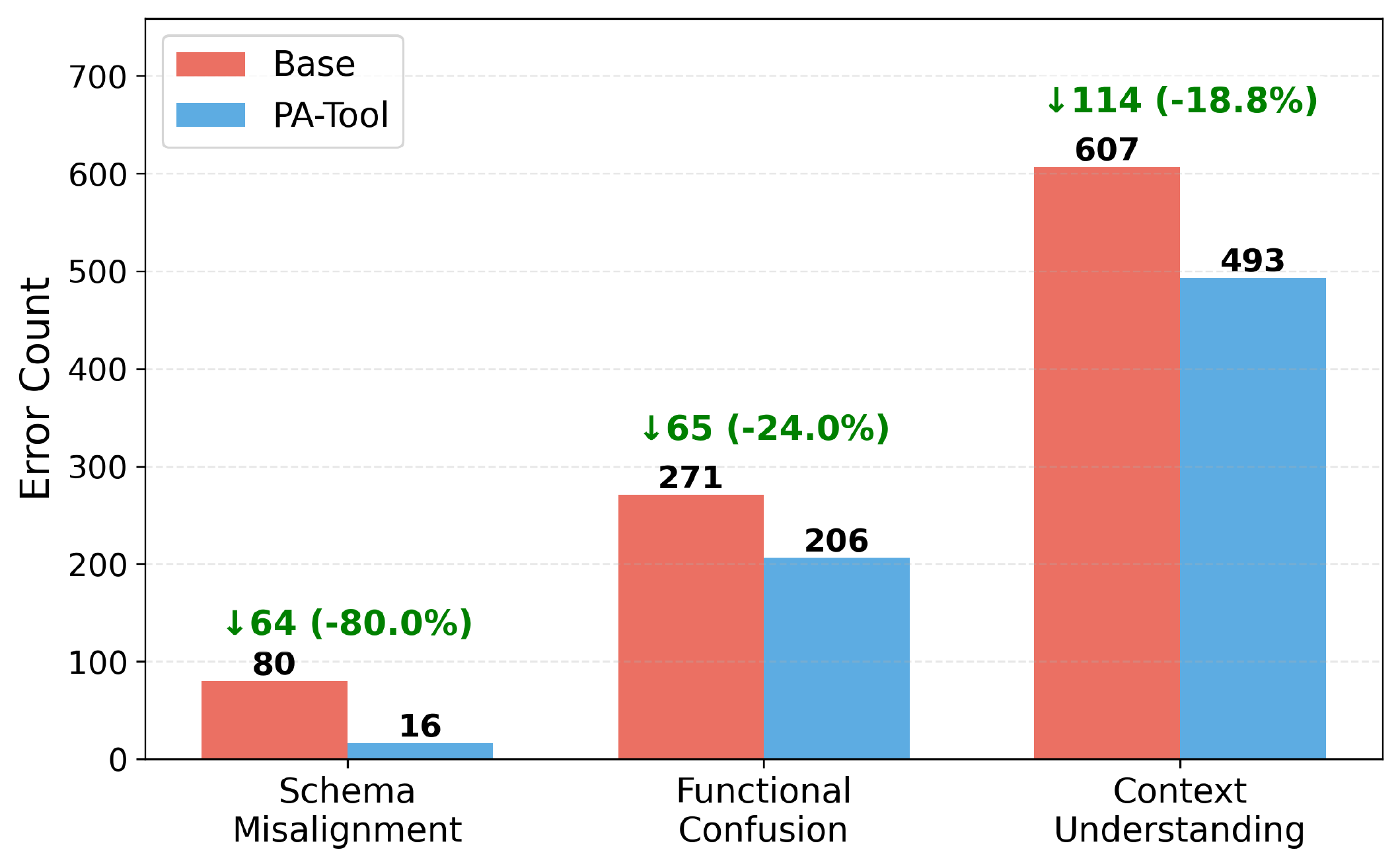}
    \caption{Error count distribution for Llama3.1-8B on MetaTool tool selection tasks.}
    \label{fig:error}
    \vspace{-1em}
\end{figure}

Figure~\ref{fig:error} shows how PA-Tool affects different error types. PA-Tool substantially reduces all error types, with particularly strong impact on Schema Misalignment (80.0\% reduction) and meaningful improvements on Functional Confusion (24.0\%) and Context Understanding (18.8\%). This demonstrates that schema alignment not only reduces naming-related failures but also indirectly benefits other aspects of tool selection, suggesting that misaligned names contribute to other error sources.

% =================================================\

% \subsection{Validating Peakedness}\label{sec:peakedness_validation}
% PA-Tool assumes that peakedness reflects models' familiarity with naming patterns. To validate this, we simulate pretraining conditions where models encounter tool schemas and measure whether peakedness increases across training epochs. 

\begin{figure}[!t]
    \centering
    \includegraphics[width=0.98\linewidth]{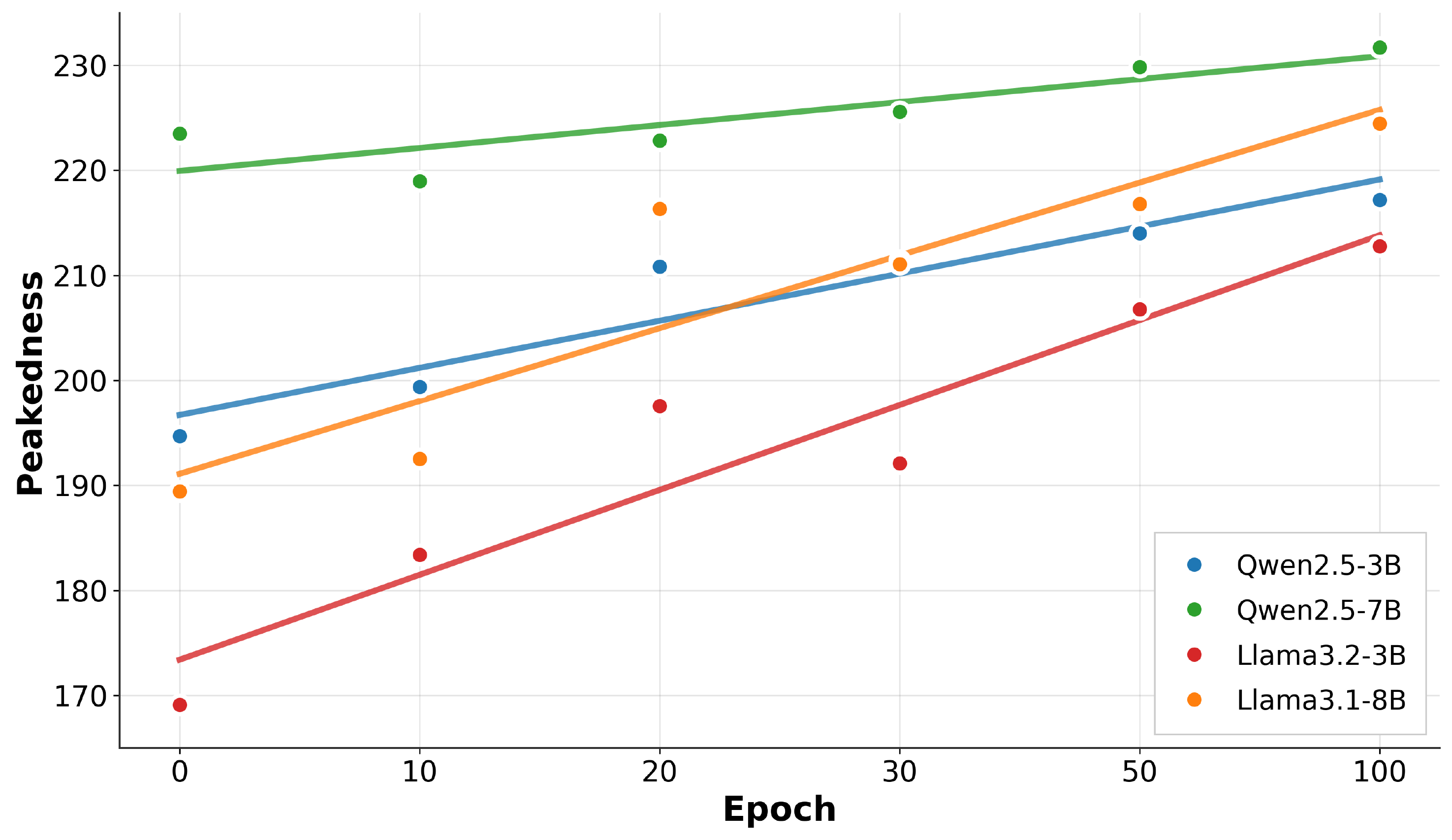}
    \caption{Peakedness across training epochs.}
    \label{fig:validation}
\end{figure}

% We use the same MetaTool training data and settings from Section~\ref{sec:sft} but with a different loss computation: unlike SFT, which computes loss only on assistant responses, we compute loss on all tokens to simulate pretraining conditions. We train four models (Qwen2.5-3B/7B, Llama3.2-3B, Llama3.1-8B), measuring peakedness at epochs 0, 10, 20, 30, 50, and 100 using 256 candidates per tool component to ensure stable measurements.

% Figure~\ref{fig:validation} shows that peakedness consistently increases with training epochs across all models, with gains up to +25.8\% (Llama3.2-3B). This supports our hypothesis that peakedness increases with training exposure, consistent with its use in PA-Tool as a familiarity signal.

\subsection{Validating Peakedness}\label{sec:peakedness_validation}
PA-Tool assumes that peakedness reflects models' familiarity 
with naming patterns. To validate this, we simulate pretraining 
conditions where models are repeatedly exposed to tool schemas 
and measure whether peakedness increases accordingly.

We use the same MetaTool data and settings from 
Section~\ref{sec:sft}, which contains user queries paired 
with tool schemas (names and descriptions). The key difference 
from SFT is the loss computation: we compute loss on all 
tokens rather than 
only on assistant responses, to simulate pretraining-style 
exposure. We train four models (Qwen2.5-3B/7B, Llama3.2-3B, 
Llama3.1-8B), measuring peakedness at epochs 0, 10, 20, 30, 
50, and 100 using 256 candidates per tool component for stable 
measurements. Figure~\ref{fig:validation} shows that peakedness consistently 
increases with training epochs across all models, with gains 
up to +25.8\% (Llama3.2-3B). This supports our hypothesis 
that peakedness increases with training exposure, consistent 
with its use in PA-Tool as a familiarity signal.

% =================================================\

\subsection{Component Ablation}\label{sec:component_ablation}
To isolate the contributions of tool name and parameter name alignment, we conduct an ablation on RoTBench, which evaluates both capabilities.

\begin{table}[t]
\centering
\resizebox{\columnwidth}{!}{
\begin{tabular}{llcccc}
\toprule
\multirow{2}{*}{\textbf{Model}} & \multirow{2}{*}{\textbf{Config.}} & \multicolumn{2}{c}{\textbf{Single-turn}} & \multicolumn{2}{c}{\textbf{Multi-turn}} \\
\cmidrule(lr){3-4} \cmidrule(lr){5-6}
& & \textbf{Tool Sel.} & \textbf{Param Id.} & \textbf{Tool Sel.} & \textbf{Param Id.} \\
\midrule
\multirow{4}{*}{Llama3.2-3B}
& Base & 56.2 & 20.0 & 32.9 & 27.1 \\
& Tool-only & 57.1 & 21.9 & \textbf{35.7} & 25.7 \\
& Param-only & 58.1 & \textbf{23.8} & 31.4 & \textbf{28.6} \\
& Both (PA-Tool) & \textbf{62.9} & 21.9 & 34.3 & \textbf{28.6}

\\
\midrule
\multirow{4}{*}{Llama3.1-8B}
& Base & 58.1 & 17.1 & 42.8 & 34.3 \\
& Tool-only & 62.9 & 14.3 & 47.1 & 34.3 \\
& Param-only & 56.2 & 17.1 & 45.7 & 34.3 \\
& Both (PA-Tool) & \textbf{68.6} & \textbf{18.1} & \textbf{48.6} & \textbf{35.7} \\

\bottomrule
\end{tabular}
}
\caption{Ablation of tool name and parameter name alignment on RoTBench.}
\label{tab:component_ablation}
\vspace{-1em}
\end{table}

Table~\ref{tab:component_ablation} shows that each alignment component tends to benefit its target capability: for example, tool name alignment improves tool selection on Llama3.1-8B (single-turn: 58.1$\rightarrow$62.9\%), while parameter name alignment improves parameter identification on Llama3.2-3B (single-turn: 20.0$\rightarrow$23.8\%). Combining both yields the most balanced results across settings: for Llama3.2-3B, it attains the highest single-turn tool selection while remaining competitive elsewhere, and for Llama3.1-8B, the joint configuration achieves the best performance across all settings. We therefore adopt the joint configuration as the default.

% =================================================
\subsection{Integration with Supervised Fine-tuning}
\label{sec:sft}
While PA-Tool is training-free, we examine how it compares to and combines with supervised fine-tuning (SFT) on MetaTool. We compare six configurations using Llama3.1-8B: (1) Base, (2) Base + PA-Tool, (3) SFT1 ($\sim$2.5K samples), (4) SFT1 + PA-Tool, (5) SFT2 (2$\times$ data, $\sim$5K samples), and (6) SFT2 + PA-Tool. Although SFT is trained on MetaTool data, 
PA-Tool is applied independently to each benchmark's own tool 
schemas. Training details and additional experiments on other models (i.e., Qwen2.5-3B/7B) are in Appendix~\ref{app:sft_setup}.

\begin{table}[t]
    \centering
    \resizebox{\columnwidth}{!}{
    \begin{tabular}{lccccc}
        \hline
        \multirow{2}{*}{\textbf{Configuration}} & \multicolumn{4}{c}{\textbf{MetaTool}} & \textbf{RoTBench} \\
        \cmidrule(lr){2-5} \cmidrule(lr){6-6}
        & \textbf{Similar} & \textbf{Scenario} & \textbf{Reliab.} & \textbf{Multi.} & \textbf{Tool Sel.} \\
        \hline
        Base & 65.2 & 71.1 & 55.1 & 79.8 & 58.1 \\
        + PA-Tool & 72.7 & 75.9 & \textbf{67.2} & \textbf{89.9} & \textbf{68.6} \\
        + SFT1 & 71.2 & 77.6 & 57.1 & 82.8 & 61.0 \\
        + SFT1 + PA-Tool & 72.7 & \textbf{80.8} & 57.1 & \textbf{89.9} & 65.7 \\
        + SFT2 (2$\times$ data) & 71.2 & 80.3 & 58.1 & 81.8 & 59.1 \\
        + SFT2 + PA-Tool & \textbf{73.2} & 79.2 & 55.6 & 86.9 & 61.9 \\
        \hline
    \end{tabular}
    }
    \caption{Results of fine-tuning and PA-Tool on 20\% random sample from MetaTool and full RoTBench.}
    \label{tab:sft}
    \vspace{-1em}
\end{table}

Table~\ref{tab:sft} shows PA-Tool provides gains
in most configurations. Notably, PA-Tool alone outperforms
both SFT1 and SFT2 on Reliability and Multi-tool without any
training, and combining both achieves the best results, with
SFT1 + PA-Tool reaching 80.8\% on Scenario. That PA-Tool
remains effective even after fine-tuning suggests that SFT
teaches tool-use reasoning but does not fully resolve the
model's pretrained naming preferences, leaving room for
PA-Tool to provide additional gains. SFT2 shows
marginal gains over SFT1 on MetaTool but degrades on RoTBench
relative to SFT1 (59.1\% vs.\ 61.0\%), suggesting that
additional domain-specific training narrows the model's
generalization to unseen tools. PA-Tool avoids
this risk entirely as it requires no training. 

% =================================================\

\subsection{Integration with Training-Free Methods}\label{sec:integration_with_trainingfree}
Beyond schema renaming, we evaluate PA-Tool's synergy with other training-free approaches. We test three categories of methods: (1) \textbf{Retrieval-based correction}, which maps misaligned outputs to valid tools post-hoc using BM25~\citep{bm25} or ToolLLM~\citep{qin2024toolllm}; (2) \textbf{Constrained generation}, which enforces valid tool names through JSON schema constraints during decoding; and (3) \textbf{Description enhancement} via EasyTool~\citep{yuan-etal-2025-easytool}, which rewrites tool descriptions for clarity without modifying names. These methods address different aspects of tool-use and are compatible with PA-Tool's schema alignment approach. We evaluate both standalone and combined configurations (Method + PA-Tool), with detailed settings presented in Appendix~\ref{app:comparisonwithothermodels}.

Table~\ref{tab:postprocessing} shows that retrieval methods yield
limited gains ($<$3\% improvement), as post-hoc
matching often fails to recover the model's intended tool.
Constrained generation substantially improves RoTBench
(Qwen2.5-7B: 78.1\% vs.\ Base 49.5\%) by eliminating format
errors, but shows mixed results on MetaTool (Table~\ref{tab:constrained}). Combining PA-Tool
with retrieval or constrained generation improves
results on most subtasks (Tables~\ref{tab:postprocessing},~\ref{tab:constrained}),
confirming these approaches are complementary.

For description enhancement (Table~\ref{tab:easytool}), PA-Tool
and EasyTool show complementary strengths: PA-Tool achieves larger
gains on tool selection for Llama models (e.g., Llama3.2-3B
Similar: 65.7\% vs.\ 45.3\%), while EasyTool is more effective on
some Qwen configurations. Combining both yields the best performance on RoTBench for most models (e.g., Qwen2.5-3B single-turn: 7.6\%→19.1\%), as name alignment and description clarity address orthogonal axis of schema quality.

% =================================================\

\subsection{Additional Benchmarks}\label{sec:other_benchmarks}

To demonstrate that PA-Tool generalizes across different evaluation settings, we evaluate on API-Bank~\cite{li-etal-2023-api} and $\tau$-Bench~\cite{yao2025taubench}. API-Bank evaluates tool-use accuracy by measuring exact matches between predicted and ground-truth tools and parameters. $\tau$-Bench is an end-to-end benchmark that assesses agent performance through multi-turn dialogues where agents must complete tasks through interaction with users. Detailed settings are in Appendix~\ref{app:evaluation_protocol}.

\paragraph{API-Bank.} PA-Tool consistently improves performance in most settings (Table~\ref{tab:other_benchmarks}). Qwen2.5-3B shows substantial gains on the Call task (18.0\%→28.5\%), while improvements persist on Call+Retrieve, which requires both tool selection and information retrieval. This confirms that schema alignment enhances tool-use accuracy.

\paragraph{$\tau$-Bench.} To verify that improved tool selection translates to better task completion, we evaluate on $\tau$-Bench (Retail). We run each evaluation 5 times to mitigate variance from the user simulator. PA-Tool consistently improves end-to-end task completion across all models (Table~\ref{tab:other_benchmarks}). Analyzing Llama3.1-8B's tool-use trajectories reveals that schema misalignment errors decreased from 115 to 98 cases, demonstrating that addressing schema misalignment contributes substantially to the observed performance gains.

\begin{table}[t]
\centering
\resizebox{\columnwidth}{!}{
\begin{tabular}{lcccccc}
\toprule
& \multicolumn{2}{c}{\textbf{Call}} & \multicolumn{2}{c}{\textbf{Call+Retrieve}} & \multicolumn{2}{c}{\textbf{$\tau$-Bench}} \\
\cmidrule(lr){2-3} \cmidrule(lr){4-5} \cmidrule(lr){6-7}
\textbf{Model} & Base & PA-Tool & Base & PA-Tool & Base & PA-Tool \\
\midrule
Qwen2.5-3B & 18.0 & \textbf{28.5} & 18.5 & \textbf{21.9} & 3.7 & \textbf{3.8} \\
Qwen2.5-7B & 25.7 & \textbf{34.7} & 23.5 & \textbf{26.9} & 6.8 & \textbf{9.7} \\
Llama3.2-3B & 5.1 & \textbf{5.4} & \textbf{10.9} & 9.2 & 4.5 & \textbf{5.6} \\
Llama3.1-8B & 28.0 & \textbf{29.8} & 21.0 & \textbf{22.4} & 9.7 & \textbf{11.1} \\
\bottomrule
\end{tabular}
}
\caption{Performance on API-Bank and end-to-end task completion (\%) on $\tau$-Bench (Retail, N=5).}
\label{tab:other_benchmarks}
\end{table}

% =================================================\

\subsection{Human Evaluation of Renamed Schemas}\label{sec:human_eval}

While PA-Tool improves model performance, a natural concern is whether optimizing schemas for model familiarity produces names that are unintuitive or obscure to human developers. To address this, we conduct a human evaluation comparing PA-Tool names with the original names.
 
Three annotators with software development experience evaluated all 199 MetaTool tools. Each annotator was presented with a tool's functional description alongside the original name and the PA-Tool name (generated by Llama3.1-8B) in a blind setup (labeled ``Name A'' and ``Name B'' in randomized order). For each pair, annotators (1) rated both names on a 1--5 scale for \textit{ease of understanding} and \textit{match to functionality}, and (2) indicated which name they preferred. This yielded 597 responses.
 
\begin{table}[t]
\centering
\small
\begin{tabular}{lcc c cc}
\toprule
& \multicolumn{2}{c}{\textbf{Rating}} & & \multicolumn{2}{c}{\textbf{Preference}} \\
\cmidrule(lr){2-3} \cmidrule(lr){5-6}
\textbf{Dimension} & Orig & PA & $\Delta$ & Orig & PA \\
\midrule
Understand. & 2.72 & \textbf{3.41} & +.69 & 10.2 & \textbf{52.3} \\
Func. match & 2.78 & \textbf{3.44} & +.66 & 10.4 & \textbf{50.9} \\
\bottomrule
\end{tabular}
\caption{Human evaluation of PA-Tool names vs.\ original names (3 annotators $\times$ 199 tools).}
\label{tab:human_eval}
\vspace{-1em}
\end{table}
 
As shown in Table~\ref{tab:human_eval}, PA-Tool names receive higher ratings on both dimensions, with improvements that are statistically significant (paired $t$-test, $p < 10^{-35}$). Over 50\% of responses prefer the PA-Tool name, while only about 10\% prefer the original. Table~\ref{tab:human_eval_examples} illustrates representative cases. The largest clarity gains occur when original names are brand-specific or opaque (e.g., \texttt{Figlet}→\texttt{ascii\_converter}, $\Delta$=+3.00), while the few cases where PA-Tool names are rated lower involve originals that are already intuitive compound words (e.g., \texttt{ProductComparison}→\texttt{compare\_options}, $\Delta$=$-$1.67), where renaming sacrifices specificity. These results indicate that PA-Tool's alignment with pretrained knowledge also yields names that are more readable and functionally descriptive for human developers. PA-Tool can therefore be deployed without sacrificing human interpretability.
 
\begin{table}[t]
\centering
\small
\resizebox{\columnwidth}{!}{
\begin{tabular}{llcl}
\toprule
\textbf{Original} & \textbf{PA-Tool} & \textbf{$\Delta$} & \textbf{Description} \\
\midrule
\multicolumn{4}{l}{\textit{PA-Tool improves clarity}} \\
\texttt{Figlet} & \texttt{ascii\_converter} & +3.00 & Convert text into ASCII fonts \\
\texttt{ad4mat} & \texttt{track\_traffic} & +2.67 & Monetize traffic via tracking links \\
\texttt{universal} & \texttt{web\_analyzer} & +2.67 & Access web pages, analyze PDFs, etc. \\
\midrule
\multicolumn{4}{l}{\textit{PA-Tool reduces clarity}} \\
\texttt{ProductComparison} & \texttt{compare\_options} & $-$1.67 & Compare product options \\
\texttt{StrologyTool} & \texttt{astro\_services} & $-$1.00 & Provide astrology services \\
\texttt{ShoppingAssistant} & \texttt{cart\_qr\_generator} & $-$1.00 & Manage cart and display QR codes \\
\bottomrule
\end{tabular}
}
\caption{Representative examples of clarity changes.}
\label{tab:human_eval_examples}
\vspace{-1em}
\end{table}

% =================================================\
\subsection{Cross-Model Schema Transfer}\label{sec:cross_model_main}
In multi-agent systems where multiple SLMs handle different subtasks, generating a separate schema for each model could become burdensome. We test whether schemas aligned to one model can also benefit a different model (full results in Appendix~\ref{app:cross_model}). Cross-model schemas generally improve over unaligned baselines, even across model families (e.g., Llama3.1-8B with Qwen2.5-7B schemas gains +8.1\% on Similar; see Table~\ref{tab:cross_model}), suggesting partially overlapping naming conventions from similar training corpora. Self-generated schemas are optimal in most settings, but the gap is small on most tasks, and larger models' schemas can benefit smaller models within the same family. Given the negligible one-time cost per model (Appendix~\ref{app:computational_time}), model-specific generation remains the preferred option, though cross-model transfer provides a practical fallback when this is not possible.

%=============================================
\section{Conclusion\label{sec:conclusion}}
We introduced PA-Tool, a training-free method that aligns tool schemas with models' pretrained knowledge by leveraging peakedness as a pretraining familiarity signal. Experiments demonstrate improvements of up to 17\% with schema misalignment errors reduced by 80\%, validating schema adaptation as an effective strategy for enhancing tool use in small language models.
PA-Tool's practical advantages make it particularly valuable for resource-constrained deployments—as a simple schema-level intervention, it can be applied without model training or fine-tuning, requiring only straightforward name mapping. By bridging pretrained knowledge and real-world tool interfaces, PA-Tool unlocks small models' potential for tool-augmented applications while preserving computational efficiency.

%=============================================
\clearpage
\section*{Limitations\label{sec:limitations}}

PA-Tool primarily targets SLMs, where schema
misalignment is a dominant failure mode, and consistently improves
performance in this regime. As model capacity increases, schema
misalignment becomes less severe due to stronger reasoning
abilities, and gains naturally diminish. This reflects the reduced
prevalence of the target problem rather than a limitation of the
approach itself; even at larger scales, PA-Tool provides meaningful
improvements in settings where misalignment compounds, such as
multi-tool composition tasks (e.g., GPT-4.1-mini gains 12\% on
MetaTool Multi-tool).

Our reliance on peakedness as an alignment signal assumes this metric reliably indicates pretraining familiarity. While validated across our experiments, this relationship may vary for models with substantially different training distributions. Additionally, our evaluation focuses on English-language schemas; the effectiveness of character-level metrics may differ for non-Latin scripts or morphologically complex languages.

PA-Tool operates on tool and parameter names, leaving
descriptions unchanged. Extending the peakedness mechanism to descriptions
risks semantic drift, where the generated description may diverge
from the tool's actual functionality. Separately, PA-Tool is compatible with description enhancement approaches that operate on an orthogonal axis. As demonstrated in our EasyTool and EasyTool with PA-Tool experiments (\S\ref{sec:integration_with_trainingfree}), combining
PA-Tool with description enhancement yields complementary gains,
suggesting that integration with interaction-driven description
refinement methods~\citep{qu2025from,
wang-etal-2024-llms-imaginarium, fang-etal-2025-play2prompt} is a
promising direction that we leave to future work.

\section*{Acknowledgments\label{sec:acknowledgements}}

This work was supported by the National Research Foundation of Korea (NRF) under the grant RS-2024-00333484 and by the Institute of Information \& Communications Technology Planning \& Evaluation (IITP) under the Leading Generative AI Human Resources Development grant IITP-2026-RS-2024-00397085 and the grant RS-2025-02215122 (Development and Demonstration of Lightweight AI Model for Smart Homes), all funded by the Korean government (MSIT). This research was also supported by the Korea Institute of Science and Technology Information (KISTI) in 2026 (No. (KISTI)K26L3M1C1), aimed at developing KONI (KISTI Open Neural Intelligence), a large language model specialized in science and technology.

\bibliography{custom}

@article{qu2025toolsurvey,
  title={Tool learning with large language models: A survey},
  author={Qu, Changle and Dai, Sunhao and Wei, Xiaochi and Cai, Hengyi and Wang, Shuaiqiang and Yin, Dawei and Xu, Jun and Wen, Ji-Rong},
  journal={Frontiers of Computer Science},
  volume={19},
  number={8},
  pages={198343},
  year={2025},
  publisher={Springer}
}

@misc{belcak2025smalllanguagemodelsfuture,
  title        = {Small Language Models are the Future of Agentic AI},
  author       = {Belcak, Peter and Heinrich, Greg and Diao, Shizhe and Fu, Yonggan and Dong, Xin and Muralidharan, Saurav and Lin, Yingyan Celine and Molchanov, Pavlo},
  year         = {2025},
  eprint       = {2506.02153},
  archivePrefix= {arXiv},
  primaryClass = {cs.AI},
  url          = {https://arxiv.org/abs/2506.02153},
  doi          = {10.48550/arXiv.2506.02153}
}

@inproceedings{chen-etal-2025-octopus,
    title = "Octopus: On-device language model for function calling of software {API}s",
    author = "Chen, Wei  and
      Li, Zhiyuan  and
      Ma, Mingyuan",
    editor = "Chen, Weizhu  and
      Yang, Yi  and
      Kachuee, Mohammad  and
      Fu, Xue-Yong",
    booktitle = "Proceedings of the 2025 Conference of the Nations of the Americas Chapter of the Association for Computational Linguistics: Human Language Technologies (Volume 3: Industry Track)",
    month = apr,
    year = "2025",
    address = "Albuquerque, New Mexico",
    publisher = "Association for Computational Linguistics",
    url = "https://aclanthology.org/2025.naacl-industry.27/",
    doi = "10.18653/v1/2025.naacl-industry.27",
    pages = "329--339",
    ISBN = "979-8-89176-194-0"
}

@inproceedings{
patil2025the,
title={The Berkeley Function Calling Leaderboard ({BFCL}): From Tool Use to Agentic Evaluation of Large Language Models},
author={Shishir G Patil and Huanzhi Mao and Fanjia Yan and Charlie Cheng-Jie Ji and Vishnu Suresh and Ion Stoica and Joseph E. Gonzalez},
booktitle={Forty-second International Conference on Machine Learning},
year={2025},
url={https://openreview.net/forum?id=2GmDdhBdDk}
}

@inproceedings{erdogan-etal-2024-tinyagent,
    title = "{T}iny{A}gent: Function Calling at the Edge",
    author = "Erdogan, Lutfi Eren  and
      Lee, Nicholas  and
      Jha, Siddharth  and
      Kim, Sehoon  and
      Tabrizi, Ryan  and
      Moon, Suhong  and
      Hooper, Coleman Richard Charles  and
      Anumanchipalli, Gopala  and
      Keutzer, Kurt  and
      Gholami, Amir",
    editor = "Hernandez Farias, Delia Irazu  and
      Hope, Tom  and
      Li, Manling",
    booktitle = "Proceedings of the 2024 Conference on Empirical Methods in Natural Language Processing: System Demonstrations",
    month = nov,
    year = "2024",
    address = "Miami, Florida, USA",
    publisher = "Association for Computational Linguistics",
    url = "https://aclanthology.org/2024.emnlp-demo.9/",
    doi = "10.18653/v1/2024.emnlp-demo.9",
    pages = "80--88"
}

@inproceedings{dong-etal-2024-generalization,
    title = "Generalization or Memorization: Data Contamination and Trustworthy Evaluation for Large Language Models",
    author = "Dong, Yihong  and
      Jiang, Xue  and
      Liu, Huanyu  and
      Jin, Zhi  and
      Gu, Bin  and
      Yang, Mengfei  and
      Li, Ge",
    editor = "Ku, Lun-Wei  and
      Martins, Andre  and
      Srikumar, Vivek",
    booktitle = "Findings of the Association for Computational Linguistics: ACL 2024",
    month = aug,
    year = "2024",
    address = "Bangkok, Thailand",
    publisher = "Association for Computational Linguistics",
    url = "https://aclanthology.org/2024.findings-acl.716/",
    doi = "10.18653/v1/2024.findings-acl.716",
    pages = "12039--12050"
}

@misc{li2023estimatingcontaminationperplexityquantifying,
      title={Estimating Contamination via Perplexity: Quantifying Memorisation in Language Model Evaluation}, 
      author={Yucheng Li},
      year={2023},
      eprint={2309.10677},
      archivePrefix={arXiv},
      primaryClass={cs.CL},
      url={https://arxiv.org/abs/2309.10677}, 
}

@inproceedings{
shi2024detecting,
title={Detecting Pretraining Data from Large Language Models},
author={Weijia Shi and Anirudh Ajith and Mengzhou Xia and Yangsibo Huang and Daogao Liu and Terra Blevins and Danqi Chen and Luke Zettlemoyer},
booktitle={The Twelfth International Conference on Learning Representations},
year={2024},
url={https://openreview.net/forum?id=zWqr3MQuNs}
}

@article{DBLP:journals/corr/abs-2107-03374,
  author       = {Mark Chen and
                  Jerry Tworek and
                  Heewoo Jun and
                  Qiming Yuan and
                  Henrique Pond{\'{e}} de Oliveira Pinto and
                  Jared Kaplan and
                  Harri Edwards and
                  Yuri Burda and
                  Nicholas Joseph and
                  Greg Brockman and
                  Alex Ray and
                  Raul Puri and
                  Gretchen Krueger and
                  Michael Petrov and
                  Heidy Khlaaf and
                  Girish Sastry and
                  Pamela Mishkin and
                  Brooke Chan and
                  Scott Gray and
                  Nick Ryder and
                  Mikhail Pavlov and
                  Alethea Power and
                  Lukasz Kaiser and
                  Mohammad Bavarian and
                  Clemens Winter and
                  Philippe Tillet and
                  Felipe Petroski Such and
                  Dave Cummings and
                  Matthias Plappert and
                  Fotios Chantzis and
                  Elizabeth Barnes and
                  Ariel Herbert{-}Voss and
                  William Hebgen Guss and
                  Alex Nichol and
                  Alex Paino and
                  Nikolas Tezak and
                  Jie Tang and
                  Igor Babuschkin and
                  Suchir Balaji and
                  Shantanu Jain and
                  William Saunders and
                  Christopher Hesse and
                  Andrew N. Carr and
                  Jan Leike and
                  Joshua Achiam and
                  Vedant Misra and
                  Evan Morikawa and
                  Alec Radford and
                  Matthew Knight and
                  Miles Brundage and
                  Mira Murati and
                  Katie Mayer and
                  Peter Welinder and
                  Bob McGrew and
                  Dario Amodei and
                  Sam McCandlish and
                  Ilya Sutskever and
                  Wojciech Zaremba},
  title        = {Evaluating Large Language Models Trained on Code},
  journal      = {CoRR},
  volume       = {abs/2107.03374},
  year         = {2021},
  url          = {https://arxiv.org/abs/2107.03374},
  eprinttype   = {arXiv},
  eprint       = {2107.03374},
  bibsource    = {dblp computer science bibliography, https://dblp.org}
}

@article{Levenshtein1965BinaryCC,
  title={Binary codes capable of correcting deletions, insertions, and reversals},
  author={Vladimir I. Levenshtein},
  journal={Soviet physics. Doklady},
  year={1965},
  volume={10},
  pages={707-710},
  url={https://api.semanticscholar.org/CorpusID:60827152}
}

@inproceedings{
shinn2023reflexion,
title={Reflexion: language agents with verbal reinforcement learning},
author={Noah Shinn and Federico Cassano and Ashwin Gopinath and Karthik R Narasimhan and Shunyu Yao},
booktitle={Thirty-seventh Conference on Neural Information Processing Systems},
year={2023},
url={https://openreview.net/forum?id=vAElhFcKW6}
}

@inproceedings{
agashe2025agent,
title={Agent S: An Open Agentic Framework that Uses Computers Like a Human},
author={Saaket Agashe and Jiuzhou Han and Shuyu Gan and Jiachen Yang and Ang Li and Xin Eric Wang},
booktitle={The Thirteenth International Conference on Learning Representations},
year={2025},
url={https://openreview.net/forum?id=lIVRgt4nLv}
}

@article{Sapkota_2026,
   title={AI Agents vs. Agentic AI: A Conceptual taxonomy, applications and challenges},
   volume={126},
   ISSN={1566-2535},
   url={http://dx.doi.org/10.1016/j.inffus.2025.103599},
   DOI={10.1016/j.inffus.2025.103599},
   journal={Information Fusion},
   publisher={Elsevier BV},
   author={Sapkota, Ranjan and Roumeliotis, Konstantinos I. and Karkee, Manoj},
   year={2026},
   month=feb, pages={103599} }

@inproceedings{cheng-etal-2024-small,
    title = "Small Agent Can Also Rock! Empowering Small Language Models as Hallucination Detector",
    author = "Cheng, Xiaoxue  and
      Li, Junyi  and
      Zhao, Xin  and
      Zhang, Hongzhi  and
      Zhang, Fuzheng  and
      Zhang, Di  and
      Gai, Kun  and
      Wen, Ji-Rong",
    editor = "Al-Onaizan, Yaser  and
      Bansal, Mohit  and
      Chen, Yun-Nung",
    booktitle = "Proceedings of the 2024 Conference on Empirical Methods in Natural Language Processing",
    month = nov,
    year = "2024",
    address = "Miami, Florida, USA",
    publisher = "Association for Computational Linguistics",
    url = "https://aclanthology.org/2024.emnlp-main.809/",
    doi = "10.18653/v1/2024.emnlp-main.809",
    pages = "14600--14615"
}

@article{schick2023toolformer,
  title={Toolformer: Language models can teach themselves to use tools},
  author={Schick, Timo and Dwivedi-Yu, Jane and Dess{\`\i}, Roberto and Raileanu, Roberta and Lomeli, Maria and Hambro, Eric and Zettlemoyer, Luke and Cancedda, Nicola and Scialom, Thomas},
  journal={Advances in Neural Information Processing Systems},
  volume={36},
  pages={68539--68551},
  year={2023}
}

@article{hsieh2023tool,
  title={Tool documentation enables zero-shot tool-usage with large language models},
  author={Hsieh, Cheng-Yu and Chen, Si-An and Li, Chun-Liang and Fujii, Yasuhisa and Ratner, Alexander and Lee, Chen-Yu and Krishna, Ranjay and Pfister, Tomas},
  journal={arXiv preprint arXiv:2308.00675},
  year={2023}
}

@inproceedings{
huang2024metatool,
title={MetaTool Benchmark for Large Language Models: Deciding Whether to Use Tools and Which to Use},
author={Yue Huang and Jiawen Shi and Yuan Li and Chenrui Fan and Siyuan Wu and Qihui Zhang and Yixin Liu and Pan Zhou and Yao Wan and Neil Zhenqiang Gong and Lichao Sun},
booktitle={The Twelfth International Conference on Learning Representations},
year={2024},
url={https://openreview.net/forum?id=R0c2qtalgG}
}

@inproceedings{li-etal-2023-api,
    title = "{API}-Bank: A Comprehensive Benchmark for Tool-Augmented {LLM}s",
    author = "Li, Minghao  and
      Zhao, Yingxiu  and
      Yu, Bowen  and
      Song, Feifan  and
      Li, Hangyu  and
      Yu, Haiyang  and
      Li, Zhoujun  and
      Huang, Fei  and
      Li, Yongbin",
    editor = "Bouamor, Houda  and
      Pino, Juan  and
      Bali, Kalika",
    booktitle = "Proceedings of the 2023 Conference on Empirical Methods in Natural Language Processing",
    month = dec,
    year = "2023",
    address = "Singapore",
    publisher = "Association for Computational Linguistics",
    url = "https://aclanthology.org/2023.emnlp-main.187/",
    doi = "10.18653/v1/2023.emnlp-main.187",
    pages = "3102--3116"
}

@inproceedings{ye-etal-2024-rotbench,
    title = "{R}o{TB}ench: A Multi-Level Benchmark for Evaluating the Robustness of Large Language Models in Tool Learning",
    author = "Ye, Junjie  and
      Wu, Yilong  and
      Gao, Songyang  and
      Huang, Caishuang  and
      Li, Sixian  and
      Li, Guanyu  and
      Fan, Xiaoran  and
      Zhang, Qi  and
      Gui, Tao  and
      Huang, Xuanjing",
    editor = "Al-Onaizan, Yaser  and
      Bansal, Mohit  and
      Chen, Yun-Nung",
    booktitle = "Proceedings of the 2024 Conference on Empirical Methods in Natural Language Processing",
    month = nov,
    year = "2024",
    address = "Miami, Florida, USA",
    publisher = "Association for Computational Linguistics",
    url = "https://aclanthology.org/2024.emnlp-main.19/",
    doi = "10.18653/v1/2024.emnlp-main.19",
    pages = "313--333"
}

@inproceedings{chen-etal-2024-eval,
    title = "{T}-Eval: Evaluating the Tool Utilization Capability of Large Language Models Step by Step",
    author = "Chen, Zehui  and
      Du, Weihua  and
      Zhang, Wenwei  and
      Liu, Kuikun  and
      Liu, Jiangning  and
      Zheng, Miao  and
      Zhuo, Jingming  and
      Zhang, Songyang  and
      Lin, Dahua  and
      Chen, Kai  and
      Zhao, Feng",
    editor = "Ku, Lun-Wei  and
      Martins, Andre  and
      Srikumar, Vivek",
    booktitle = "Proceedings of the 62nd Annual Meeting of the Association for Computational Linguistics (Volume 1: Long Papers)",
    month = aug,
    year = "2024",
    address = "Bangkok, Thailand",
    publisher = "Association for Computational Linguistics",
    url = "https://aclanthology.org/2024.acl-long.515/",
    doi = "10.18653/v1/2024.acl-long.515",
    pages = "9510--9529"
}

@inproceedings{
seo2025simuhometemporalenvironmentawarebenchmark,
title={SimuHome: A Temporal- and Environment-Aware Benchmark for Smart Home {LLM} Agents},
author={Gyuhyeon Seo and Jungwoo Yang and Junseong Pyo and Nalim Kim and Jonggeun Lee and Yohan Jo},
booktitle={The Fourteenth International Conference on Learning Representations},
year={2026},
url={https://openreview.net/forum?id=LCS1WsGvha}
}

@inproceedings{
yao2025taubench,
title={\{\${\textbackslash}tau\$\}-bench: A Benchmark for {\textbackslash}underline\{T\}ool-{\textbackslash}underline\{A\}gent-{\textbackslash}underline\{U\}ser Interaction in Real-World Domains},
author={Shunyu Yao and Noah Shinn and Pedram Razavi and Karthik R Narasimhan},
booktitle={The Thirteenth International Conference on Learning Representations},
year={2025},
url={https://openreview.net/forum?id=roNSXZpUDN}
}

@inproceedings{trivedi-etal-2024-appworld,
    title = "{A}pp{W}orld: A Controllable World of Apps and People for Benchmarking Interactive Coding Agents",
    author = "Trivedi, Harsh  and
      Khot, Tushar  and
      Hartmann, Mareike  and
      Manku, Ruskin  and
      Dong, Vinty  and
      Li, Edward  and
      Gupta, Shashank  and
      Sabharwal, Ashish  and
      Balasubramanian, Niranjan",
    editor = "Ku, Lun-Wei  and
      Martins, Andre  and
      Srikumar, Vivek",
    booktitle = "Proceedings of the 62nd Annual Meeting of the Association for Computational Linguistics (Volume 1: Long Papers)",
    month = aug,
    year = "2024",
    address = "Bangkok, Thailand",
    publisher = "Association for Computational Linguistics",
    url = "https://aclanthology.org/2024.acl-long.850/",
    doi = "10.18653/v1/2024.acl-long.850",
    pages = "16022--16076"
}

@inproceedings{
qin2024toolllm,
title={Tool{LLM}: Facilitating Large Language Models to Master 16000+ Real-world {API}s},
author={Yujia Qin and Shihao Liang and Yining Ye and Kunlun Zhu and Lan Yan and Yaxi Lu and Yankai Lin and Xin Cong and Xiangru Tang and Bill Qian and Sihan Zhao and Lauren Hong and Runchu Tian and Ruobing Xie and Jie Zhou and Mark Gerstein and dahai li and Zhiyuan Liu and Maosong Sun},
booktitle={The Twelfth International Conference on Learning Representations},
year={2024},
url={https://openreview.net/forum?id=dHng2O0Jjr}
}

@inproceedings{
shim2025noncollaborativeusersimulatorstool,
title={Non-Collaborative User Simulators for Tool Agents},
author={Jeonghoon Shim and Woojung Song and Cheyon Jin and Seungwon Kook and Yohan Jo},
booktitle={The Fourteenth International Conference on Learning Representations},
year={2026},
url={https://openreview.net/forum?id=UAUimofy3W}
}

@inproceedings{
liu2025toolace,
title={Tool{ACE}: Winning the Points of {LLM} Function Calling},
author={Weiwen Liu and Xu Huang and Xingshan Zeng and xinlong hao and Shuai Yu and Dexun Li and Shuai Wang and Weinan Gan and Zhengying Liu and Yuanqing Yu and Zezhong WANG and Yuxian Wang and Wu Ning and Yutai Hou and Bin Wang and Chuhan Wu and Wang Xinzhi and Yong Liu and Yasheng Wang and Duyu Tang and Dandan Tu and Lifeng Shang and Xin Jiang and Ruiming Tang and Defu Lian and Qun Liu and Enhong Chen},
booktitle={The Thirteenth International Conference on Learning Representations},
year={2025},
url={https://openreview.net/forum?id=8EB8k6DdCU}
}

@inproceedings{zhang-etal-2025-xlam,
    title = "x{LAM}: A Family of Large Action Models to Empower {AI} Agent Systems",
    author = "Zhang, Jianguo  and
      Lan, Tian  and
      Zhu, Ming  and
      Liu, Zuxin  and
      Hoang, Thai Quoc  and
      Kokane, Shirley  and
      Yao, Weiran  and
      Tan, Juntao  and
      Prabhakar, Akshara  and
      Chen, Haolin  and
      Liu, Zhiwei  and
      Feng, Yihao  and
      Awalgaonkar, Tulika Manoj  and
      R N, Rithesh  and
      Chen, Zeyuan  and
      Xu, Ran  and
      Niebles, Juan Carlos  and
      Heinecke, Shelby  and
      Wang, Huan  and
      Savarese, Silvio  and
      Xiong, Caiming",
    editor = "Chiruzzo, Luis  and
      Ritter, Alan  and
      Wang, Lu",
    booktitle = "Proceedings of the 2025 Conference of the Nations of the Americas Chapter of the Association for Computational Linguistics: Human Language Technologies (Volume 1: Long Papers)",
    month = apr,
    year = "2025",
    address = "Albuquerque, New Mexico",
    publisher = "Association for Computational Linguistics",
    url = "https://aclanthology.org/2025.naacl-long.578/",
    doi = "10.18653/v1/2025.naacl-long.578",
    pages = "11583--11597",
    ISBN = "979-8-89176-189-6"
}

@inproceedings{shi-etal-2024-direct,
    title = "Direct Multi-Turn Preference Optimization for Language Agents",
    author = "Shi, Wentao  and
      Yuan, Mengqi  and
      Wu, Junkang  and
      Wang, Qifan  and
      Feng, Fuli",
    editor = "Al-Onaizan, Yaser  and
      Bansal, Mohit  and
      Chen, Yun-Nung",
    booktitle = "Proceedings of the 2024 Conference on Empirical Methods in Natural Language Processing",
    month = nov,
    year = "2024",
    address = "Miami, Florida, USA",
    publisher = "Association for Computational Linguistics",
    url = "https://aclanthology.org/2024.emnlp-main.138/",
    doi = "10.18653/v1/2024.emnlp-main.138",
    pages = "2312--2324"
}

@inproceedings{
qian2025toolrl,
title={Tool{RL}: Reward is All Tool Learning Needs},
author={Cheng Qian and Emre Can Acikgoz and Qi He and Hongru WANG and Xiusi Chen and Dilek Hakkani-T{\"u}r and Gokhan Tur and Heng Ji},
booktitle={The Thirty-ninth Annual Conference on Neural Information Processing Systems},
year={2025},
url={https://openreview.net/forum?id=eOLdGbXT6t}
}

@inproceedings{
feng2026retool,
title={ReTool: Reinforcement Learning for Strategic Tool Use in {LLM}s},
author={Jiazhan Feng and Shijue Huang and Xingwei Qu and Ge Zhang and Yujia Qin and Baoquan Zhong and Chengquan Jiang and Jinxin Chi and Wanjun Zhong},
booktitle={The Fourteenth International Conference on Learning Representations},
year={2026},
url={https://openreview.net/forum?id=tRk1nofSmz}
}

@misc{chen2025reinforcementlearninglonghorizoninteractive,
      title={Reinforcement Learning for Long-Horizon Interactive LLM Agents}, 
      author={Kevin Chen and Marco Cusumano-Towner and Brody Huval and Aleksei Petrenko and Jackson Hamburger and Vladlen Koltun and Philipp Krähenbühl},
      year={2025},
      eprint={2502.01600},
      archivePrefix={arXiv},
      primaryClass={cs.LG},
      url={https://arxiv.org/abs/2502.01600}, 
}

@inproceedings{
fu2024autoguide,
title={AutoGuide: Automated Generation and Selection of Context-Aware Guidelines for Large Language Model Agents},
author={Yao Fu and Dong-Ki Kim and Jaekyeom Kim and Sungryull Sohn and Lajanugen Logeswaran and Kyunghoon Bae and Honglak Lee},
booktitle={The Thirty-eighth Annual Conference on Neural Information Processing Systems},
year={2024},
url={https://openreview.net/forum?id=mRIQz8Zd6O}
}

@inproceedings{wang-etal-2024-llms-imaginarium,
    title = "{LLM}s in the Imaginarium: Tool Learning through Simulated Trial and Error",
    author = "Wang, Boshi  and
      Fang, Hao  and
      Eisner, Jason  and
      Van Durme, Benjamin  and
      Su, Yu",
    editor = "Ku, Lun-Wei  and
      Martins, Andre  and
      Srikumar, Vivek",
    booktitle = "Proceedings of the 62nd Annual Meeting of the Association for Computational Linguistics (Volume 1: Long Papers)",
    month = aug,
    year = "2024",
    address = "Bangkok, Thailand",
    publisher = "Association for Computational Linguistics",
    url = "https://aclanthology.org/2024.acl-long.570/",
    doi = "10.18653/v1/2024.acl-long.570",
    pages = "10583--10604"
}

@inproceedings{zhao2024expel,
  title={Expel: Llm agents are experiential learners},
  author={Zhao, Andrew and Huang, Daniel and Xu, Quentin and Lin, Matthieu and Liu, Yong-Jin and Huang, Gao},
  booktitle={Proceedings of the AAAI Conference on Artificial Intelligence},
  volume={38},
  number={17},
  pages={19632--19642},
  year={2024}
}

@inproceedings{cui-etal-2025-enhancing-tool,
    title = "Enhancing Tool Learning in Large Language Models with Hierarchical Error Checklists",
    author = "Cui, Yue  and
      Yao, Liuyi  and
      Tao, Shuchang  and
      Shi, Weijie  and
      Li, Yaliang  and
      Ding, Bolin  and
      Zhou, Xiaofang",
    editor = "Che, Wanxiang  and
      Nabende, Joyce  and
      Shutova, Ekaterina  and
      Pilehvar, Mohammad Taher",
    booktitle = "Findings of the Association for Computational Linguistics: ACL 2025",
    month = jul,
    year = "2025",
    address = "Vienna, Austria",
    publisher = "Association for Computational Linguistics",
    url = "https://aclanthology.org/2025.findings-acl.841/",
    doi = "10.18653/v1/2025.findings-acl.841",
    pages = "16357--16375",
    ISBN = "979-8-89176-256-5"
}

@inproceedings{NEURIPS2020_1457c0d6,
 author = {Brown, Tom and Mann, Benjamin and Ryder, Nick and Subbiah, Melanie and Kaplan, Jared D and Dhariwal, Prafulla and Neelakantan, Arvind and Shyam, Pranav and Sastry, Girish and Askell, Amanda and Agarwal, Sandhini and Herbert-Voss, Ariel and Krueger, Gretchen and Henighan, Tom and Child, Rewon and Ramesh, Aditya and Ziegler, Daniel and Wu, Jeffrey and Winter, Clemens and Hesse, Chris and Chen, Mark and Sigler, Eric and Litwin, Mateusz and Gray, Scott and Chess, Benjamin and Clark, Jack and Berner, Christopher and McCandlish, Sam and Radford, Alec and Sutskever, Ilya and Amodei, Dario},
 booktitle = {Advances in Neural Information Processing Systems},
 editor = {H. Larochelle and M. Ranzato and R. Hadsell and M.F. Balcan and H. Lin},
 pages = {1877--1901},
 publisher = {Curran Associates, Inc.},
 title = {Language Models are Few-Shot Learners},
 url = {https://proceedings.neurips.cc/paper_files/paper/2020/file/1457c0d6bfcb4967418bfb8ac142f64a-Paper.pdf},
 volume = {33},
 year = {2020}
}

@misc{yang2023rethinkingbenchmarkcontaminationlanguage,
      title={Rethinking Benchmark and Contamination for Language Models with Rephrased Samples}, 
      author={Shuo Yang and Wei-Lin Chiang and Lianmin Zheng and Joseph E. Gonzalez and Ion Stoica},
      year={2023},
      eprint={2311.04850},
      archivePrefix={arXiv},
      primaryClass={cs.CL},
      url={https://arxiv.org/abs/2311.04850}, 
}

@inproceedings{yuan-etal-2025-easytool,
    title = "{EASYTOOL}: Enhancing {LLM}-based Agents with Concise Tool Instruction",
    author = "Yuan, Siyu  and
      Song, Kaitao  and
      Chen, Jiangjie  and
      Tan, Xu  and
      Shen, Yongliang  and
      Ren, Kan  and
      Li, Dongsheng  and
      Yang, Deqing",
    editor = "Chiruzzo, Luis  and
      Ritter, Alan  and
      Wang, Lu",
    booktitle = "Proceedings of the 2025 Conference of the Nations of the Americas Chapter of the Association for Computational Linguistics: Human Language Technologies (Volume 1: Long Papers)",
    month = apr,
    year = "2025",
    address = "Albuquerque, New Mexico",
    publisher = "Association for Computational Linguistics",
    url = "https://aclanthology.org/2025.naacl-long.44/",
    doi = "10.18653/v1/2025.naacl-long.44",
    pages = "951--972",
    ISBN = "979-8-89176-189-6"
}

@inproceedings{
qu2025from,
title={From Exploration to Mastery: Enabling {LLM}s to Master Tools via Self-Driven Interactions},
author={Changle Qu and Sunhao Dai and Xiaochi Wei and Hengyi Cai and Shuaiqiang Wang and Dawei Yin and Jun Xu and Ji-Rong Wen},
booktitle={The Thirteenth International Conference on Learning Representations},
year={2025},
url={https://openreview.net/forum?id=QKBu1BOAwd}
}

@misc{qwen2025qwen25technicalreport,
      title={Qwen2.5 Technical Report}, 
      author={Qwen and : and An Yang and Baosong Yang and Beichen Zhang and Binyuan Hui and Bo Zheng and Bowen Yu and Chengyuan Li and Dayiheng Liu and Fei Huang and Haoran Wei and Huan Lin and Jian Yang and Jianhong Tu and Jianwei Zhang and Jianxin Yang and Jiaxi Yang and Jingren Zhou and Junyang Lin and Kai Dang and Keming Lu and Keqin Bao and Kexin Yang and Le Yu and Mei Li and Mingfeng Xue and Pei Zhang and Qin Zhu and Rui Men and Runji Lin and Tianhao Li and Tianyi Tang and Tingyu Xia and Xingzhang Ren and Xuancheng Ren and Yang Fan and Yang Su and Yichang Zhang and Yu Wan and Yuqiong Liu and Zeyu Cui and Zhenru Zhang and Zihan Qiu},
      year={2025},
      eprint={2412.15115},
      archivePrefix={arXiv},
      primaryClass={cs.CL},
      url={https://arxiv.org/abs/2412.15115}, 
}

@misc{openai_gpt41mini_2024,
  title        = {GPT-4.1 mini},
  author       = {{OpenAI}},
  year         = {2025},
  howpublished = {\url{https://openai.com/index/gpt-4-1/}},
  note         = {Accessed 2026-04-19}
}

@misc{google_gemini25flash_2025,
  title        = {Continuing to bring you our latest models, with an improved Gemini 2.5 Flash and Flash-Lite release},
  author       = {Basu Mallick, Shrestha and Lall, Sid and Gleicher, Zach and Olszewska, Kate},
  year         = {2025},
  month        = {September},
  howpublished = {\url{https://developers.googleblog.com/en/continuing-to-bring-you-our-latest-models-with-an-improved-gemini-2-5-flash-and-flash-lite-release/}},
  note         = {Google Developers Blog; Accessed 2026-04-19}
}

@misc{meta_llama32_2024,
  title        = {Llama 3.2: Revolutionizing edge AI and vision with open, efficient models},
  author       = {{Meta AI}},
  year         = {2024},
  month        = {September},
  howpublished = {\url{https://ai.meta.com/blog/llama-3-2-connect-2024-vision-edge-mobile-devices/}},
  note         = {Accessed 2026-04-19}
}

@misc{meta_llama31_2024,
  title        = {Introducing Llama 3.1: Our most capable models to date},
  author       = {{Meta AI}},
  year         = {2024},
  howpublished = {\url{https://ai.meta.com/blog/meta-llama-3-1/}},
  note         = {Accessed 2026-04-19}
}

@misc{anthropic_2025_sonnet,
  title        = {Introducing Claude Sonnet 4.5},
  author       = {{Anthropic}},
  year         = {2025},
  howpublished = {\url{https://www.anthropic.com/news/claude-sonnet-4-5}},
  note         = {Accessed 2026-04-19}
}

@article{bm25,
author = {Robertson, Stephen and Zaragoza, Hugo},
title = {The Probabilistic Relevance Framework: BM25 and Beyond},
year = {2009},
issue_date = {April 2009},
publisher = {Now Publishers Inc.},
address = {Hanover, MA, USA},
volume = {3},
number = {4},
issn = {1554-0669},
url = {https://doi.org/10.1561/1500000019},
doi = {10.1561/1500000019},
journal = {Found. Trends Inf. Retr.},
month = apr,
pages = {333–389},
numpages = {57}
}

@inproceedings{fang-etal-2025-play2prompt,
    title = "{PLAY}2{PROMPT}: Zero-shot Tool Instruction Optimization for {LLM} Agents via Tool Play",
    author = "Fang, Wei  and
      Zhang, Yang  and
      Qian, Kaizhi  and
      Glass, James R.  and
      Zhu, Yada",
    editor = "Che, Wanxiang  and
      Nabende, Joyce  and
      Shutova, Ekaterina  and
      Pilehvar, Mohammad Taher",
    booktitle = "Findings of the Association for Computational Linguistics: ACL 2025",
    month = jul,
    year = "2025",
    address = "Vienna, Austria",
    publisher = "Association for Computational Linguistics",
    url = "https://aclanthology.org/2025.findings-acl.1347/",
    doi = "10.18653/v1/2025.findings-acl.1347",
    pages = "26274--26290",
    ISBN = "979-8-89176-256-5"
}
%=============================================
\newpage
\appendix

\section{PA-Tool}
\subsection{Detailed Algorithm for PA-Tool}\label{app:patool_algorithm}
\begin{algorithm}[H]
{\small
\caption{Pseudocode for PA-Tool}
\label{alg:patool}
\begin{algorithmic}[1]
\Require Language model $\mathcal{M}$, Component description $d$, samples $N$, temperature $t$, hyperparameter $\alpha$
\Ensure Representative name $s^*$
\State \textbf{// Stage 1: Candidate Generation}
\State $s_{\text{ref}} \gets$ Generate name from $\mathcal{M}$ with temperature 0
\State $\mathcal{C} \gets \{\}$
\For{$i = 1$ \textbf{to} $N$}
    \State $s_i \gets$ Generate name from $\mathcal{M}$ with temperature $t$
    \State Add $s_i$ to $\mathcal{C}$
\EndFor
\State \textbf{// Stage 2: Peakedness Computation}
\State $\ell_{\max} \gets$ maximum character length of name in $\mathcal{C}$
\State $\tau \gets \alpha \cdot \ell_{\max}$ \Comment{Eq. (\ref{eq:threshold})}
\For{\textbf{each} $s_i$ in $\mathcal{C}$}
    \State $\phi(s_i) \gets$ count of $s_j \in \mathcal{C}$ with $j \neq i$ and edit distance to $s_i$ $\leq \tau$ \Comment{Eq.~\eqref{eq:peakedness}}
\EndFor
\State \textbf{// Stage 3: Schema Selection}
\State $\mathcal{C}^* \gets$ names in $\mathcal{C}$ with maximum peakedness \Comment{Eq. (\ref{eq:selection})}
\If{$|\mathcal{C}^*$| == 1}
    \State $s^* \gets$ the unique name in $\mathcal{C}^*$
\Else
    \State $s^* \gets$ name in $\mathcal{C}^*$ closest to $s_{\text{ref}}$ \Comment{Eq. (\ref{eq:tiebreak})}
\EndIf
\State \textbf{return} $s^*$
\end{algorithmic}
}
\end{algorithm}

%=============================================

\subsection{Example of Schema Generation Process}
\label{app:schema_example}

We provide a concrete example of how PA-Tool generates pretraining-aligned schemas. Given the original tool name \texttt{DietTool} with description "A tool that simplifies calorie counting, tracks diet, and provides insights from many restaurants and grocery stores...", PA-Tool generates 32 candidate names by sampling at temperature 0.4.

The top candidates by frequency are:
\begin{itemize}
    \item \texttt{diet\_tracker}: 5 occurrences
    \item \texttt{diet\_insights}: 4 occurrences  
    \item \texttt{calorie\_tracker}: 3 occurrences
    \item \texttt{nutri\_guide}: 3 occurrences
    \item \texttt{eatwise}: 3 occurrences
    \item Others: \texttt{nutrify}, \texttt{nutri\_navigator}, \texttt{diet\_planner}, etc. (1-2 occurrences each)
\end{itemize}

Importantly, PA-Tool does not simply select the most frequent candidate. Instead, it computes peakedness by measuring how many similar candidates cluster around each option using edit distance. In this case, PA-Tool selects \texttt{diet\_insights} (peakedness=4) rather than the most frequent \texttt{diet\_tracker} (5 occurrences), as the former has a tighter cluster of similar variants indicating stronger distributional concentration.

It is also worth noting that greedy decoding (temperature 0) produces \texttt{nutri\_guide}, which differs from both the most frequent candidate and PA-Tool's selection. This illustrates three distinct outcomes: (1) PA-Tool's peakedness-based selection (\texttt{diet\_insights}), (2) the most frequent candidate (\texttt{diet\_tracker}), and (3) greedy decoding's output (\texttt{nutri\_guide}). These differences highlight that PA-Tool's selection mechanism considers distributional concentration rather than simple frequency or single-sample generation.

\subsection{Name Collision Resolution}\label{app:name_collision}

When tools have highly similar descriptions, PA-Tool may generate identical names for different components. We resolve this through iterative priority-based locking: in each round, the component with the highest peakedness for a contested name acquires it, while others cascade to their next candidate. This process repeats until all components have unique names.

\begin{table}[h]
\centering
\small
\begin{tabular}{lcc}
\toprule
\textbf{Model} & \textbf{MetaTool} & \textbf{RoTBench} \\
\midrule
Qwen2.5-3B & 2 (1.0\%) & 13 (4.2\%) \\
Qwen2.5-7B & 0 (0.0\%) & 15 (4.8\%) \\
Llama3.2-3B & 1 (0.5\%) & 18 (5.8\%) \\
Llama3.1-8B & 1 (0.5\%) & 17 (5.5\%) \\
\bottomrule
\end{tabular}
\caption{Name collision statistics across benchmarks.}
\label{tab:collision}
\end{table}

Table~\ref{tab:collision} shows collision frequencies across benchmarks. Collisions remain rare across all models (less than 6\%), with MetaTool showing particularly low rates (0-1.0\%) and RoTBench showing slightly higher but still modest rates (4.2-5.8\%). RoTBench's higher collision rate stems from greater lexical overlap among tool descriptions (average Jaccard similarity of 0.025 vs. 0.011 for MetaTool).

\subsection{Effect of Hyperparameters}\label{app:hyperparameter}
\begin{figure}[t]
    \centering
    \includegraphics[width=0.48\textwidth]{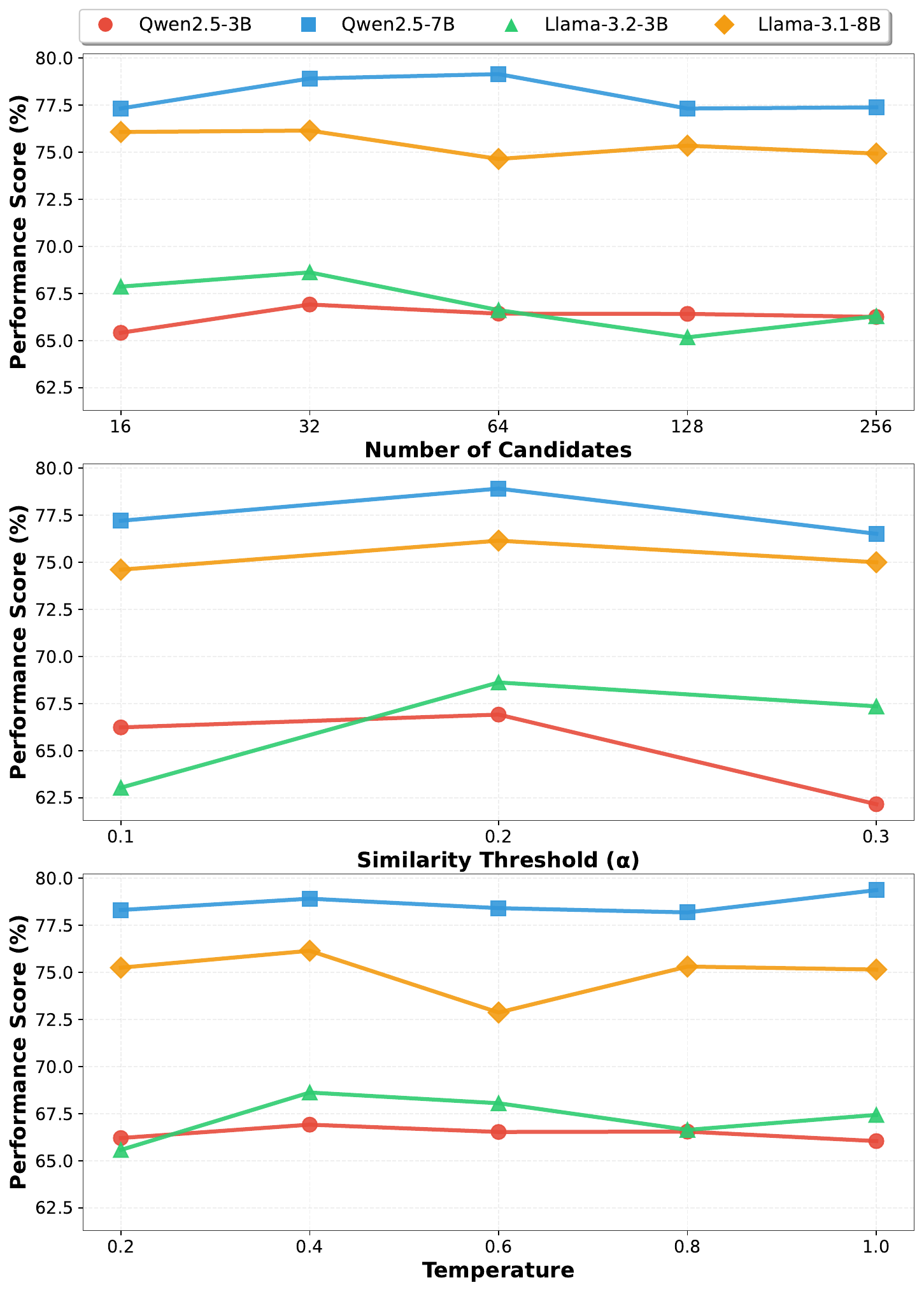}
    \caption{Effect of hyperparameters on PA-Tool across different models. All results are averaged across four MetaTool subtasks. \textbf{Top:} Effect of the number of candidates ($N$). \textbf{Middle:} Effect of similarity threshold ($\alpha$). \textbf{Bottom:} Effect of sampling temperature ($t$).}
    \label{fig:hyperparameter_analysis}
\end{figure}

\begin{table}[!h]
\centering
\resizebox{\columnwidth}{!}{
\begin{tabular}{lc|cccc}
\toprule
\multirow{2}{*}{\textbf{Model}} & \multirow{2}{*}{\textbf{N}} & \multicolumn{4}{c}{\textbf{MetaTool}} \\
\cmidrule(lr){3-6}
& & Similar & Scenario & Reliab. & Multi. \\
\midrule
\multirow{5}{*}{Qwen2.5-3B} & 16 & 50.5 & 59.1 & 85.7 & 66.4 \\
& 32 & 50.0 & 58.8 & \textbf{86.2} & \textbf{72.6} \\
& 64 & \textbf{50.9} & \textbf{59.1} & 86.1 & 69.6 \\
& 128 & 50.5 & 58.6 & 86.0 & 70.6 \\
& 256 & 50.2 & \textbf{59.1} & 83.2 & 72.4 \\
\midrule
\multirow{5}{*}{Qwen2.5-7B} & 16 & 63.2 & 74.9 & 86.8 & 84.3 \\
& 32 & \textbf{64.1} & \textbf{78.4} & \textbf{88.2} & 84.9 \\
& 64 & 62.9 & 77.7 & \textbf{88.2} & \textbf{87.7} \\
& 128 & 62.3 & 77.4 & 85.4 & 84.1 \\
& 256 & 63.0 & \textbf{78.4} & 81.0 & 87.1 \\
\midrule
\multirow{5}{*}{Llama3.2-3B} & 16 & 64.2 & 65.9 & 60.4 & \textbf{80.9} \\
& 32 & \textbf{65.7} & \textbf{67.7} & \textbf{60.6} & 80.5 \\
& 64 & 64.3 & 65.2 & 57.5 & 79.5 \\
& 128 & 63.7 & 65.2 & 52.5 & 79.3 \\
& 256 & 64.8 & 65.5 & 54.2 & 80.7 \\
\midrule
\multirow{5}{*}{Llama3.1-8B} & 16 & 68.9 & 79.5 & \textbf{66.9} & \textbf{88.9} \\
& 32 & \textbf{70.4} & \textbf{79.9} & 66.0 & 88.3 \\
& 64 & 68.7 & 77.8 & 64.5 & 87.5 \\
& 128 & 67.9 & 79.4 & 65.1 & \textbf{88.9} \\
& 256 & 67.6 & 79.8 & 63.9 & 88.3 \\
\bottomrule
\end{tabular}
}
\caption{Effect of Number of Candidates.}
\label{tab:candidates_effect}
\end{table}

%=============================================

\begin{table}[!h]
\centering
\resizebox{\columnwidth}{!}{
\begin{tabular}{ll|cccc}
\toprule
\multirow{2}{*}{\textbf{Model}} & \multirow{2}{*}{\boldmath{$\alpha$}} & \multicolumn{4}{c}{\textbf{MetaTool}} \\
\cmidrule(lr){3-6}
& & Similar & Scenario & Reliab. & Multi. \\
\midrule
\multirow{3}{*}{Qwen2.5-3B} & 0.1 & 50.5 & 58.3 & 84.1 & 72.0 \\
& 0.2 & 50.0 & \textbf{58.8} & \textbf{86.2} & \textbf{72.6} \\
& 0.3 & \textbf{51.0} & 58.6 & 78.9 & 60.2 \\
\midrule
\multirow{3}{*}{Qwen2.5-7B} & 0.1 & 62.7 & 76.4 & 84.8 & \textbf{84.9} \\
& 0.2 & \textbf{64.1} & \textbf{78.4} & \textbf{88.2} & \textbf{84.9} \\
& 0.3 & 63.2 & 75.8 & 84.7 & 82.3 \\
\midrule
\multirow{3}{*}{Llama3.2-3B} & 0.1 & 62.8 & 66.3 & 54.8 & 68.2 \\
& 0.2 & \textbf{65.7} & \textbf{67.7} & \textbf{60.6} & \textbf{80.5} \\
& 0.3 & 65.0 & 65.5 & 59.0 & 79.9 \\
\midrule
\multirow{3}{*}{Llama3.1-8B} & 0.1 & 68.3 & 76.9 & 65.0 & 88.1 \\
& 0.2 & \textbf{70.4} & \textbf{79.9} & \textbf{66.0} & 88.3 \\
& 0.3 & 68.1 & 78.7 & 64.4 & \textbf{88.7} \\
\bottomrule
\end{tabular}
}
\caption{Effect of Similarity Threshold.}
\label{tab:alpha_effect}
\end{table}

\begin{table}[!h]
\centering
\resizebox{\columnwidth}{!}{
\begin{tabular}{ll|cccc}
\toprule
\multirow{2}{*}{\textbf{Model}} & \multirow{2}{*}{\textbf{Temp.}} & \multicolumn{4}{c}{\textbf{MetaTool}} \\
\cmidrule(lr){3-6}
& & Similar & Scenario & Reliab. & Multi. \\
\midrule
\multirow{5}{*}{Qwen2.5-3B} & 0.2 & 50.2 & 58.5 & 85.0 & 71.0 \\
& 0.4 & 50.0 & 58.8 & \textbf{86.2} & \textbf{72.6} \\
& 0.6 & 51.5 & 58.9 & 85.5 & 70.2 \\
& 0.8 & 50.5 & 59.7 & 83.9 & 72.0 \\
& 1.0 & \textbf{51.7} & \textbf{60.2} & 85.1 & 67.2 \\
\midrule
\multirow{5}{*}{Qwen2.5-7B} & 0.2 & 62.2 & 75.3 & 88.8 & \textbf{86.9} \\
& 0.4 & \textbf{64.1} & \textbf{78.4} & 88.2 & 84.9 \\
& 0.6 & 62.3 & 75.8 & 88.6 & \textbf{86.9} \\
& 0.8 & 61.4 & 77.7 & 87.5 & 86.1 \\
& 1.0 & 63.1 & 77.4 & \textbf{90.2} & 86.7 \\
\midrule
\multirow{5}{*}{Llama3.2-3B} & 0.2 & 64.1 & 68.3 & 55.6 & 74.2 \\
& 0.4 & \textbf{65.7} & 67.7 & 60.6 & \textbf{80.5} \\
& 0.6 & 61.3 & 66.4 & \textbf{64.6} & 79.9 \\
& 0.8 & 63.5 & \textbf{68.7} & 54.1 & 80.3 \\
& 1.0 & 64.0 & 66.3 & 61.5 & 77.9 \\
\midrule
\multirow{5}{*}{Llama3.1-8B} & 0.2 & 69.5 & 79.5 & 63.5 & 88.5 \\
& 0.4 & \textbf{70.4} & \textbf{79.9} & \textbf{66.0} & 88.3 \\
& 0.6 & 68.4 & 79.8 & 62.3 & 80.9 \\
& 0.8 & 67.8 & 79.4 & 65.2 & \textbf{88.7} \\
& 1.0 & 69.5 & 76.9 & 65.6 & 88.5 \\
\bottomrule
\end{tabular}
}
\caption{Effect of Sampling Temperature.}
\label{tab:temperature_effect}
\end{table}

We investigate the effect of PA-Tool's three key hyperparameters on performance: the number of candidates ($N$), the similarity threshold ($\alpha$), and the sampling temperature ($t$). All experiments are conducted on the MetaTool benchmark. 

\paragraph{Number of Candidates ($N$).} Table~\ref{tab:candidates_effect} shows that smaller models (3B) achieve stable performance with 16--32 candidates, while larger models (7--8B) require 32--64 candidates before performance plateaus. Beyond these ranges, additional candidates provide minimal gains.

\paragraph{Similarity Threshold ($\alpha$).} Table~\ref{tab:alpha_effect} shows that performance peaks at $\alpha=0.2$ consistently across all models. At $\alpha=0.1$, performance drops by 2--3\%, while $\alpha=0.3$ shows similar degradation.

\paragraph{Sampling Temperature ($t$).} Table~\ref{tab:temperature_effect} shows that performance remains stable across temperatures $t \in [0.2, 1.0]$, varying within 1--2\%. Moderate temperatures ($t=0.4$--$0.6$) show slightly better results, though the differences are marginal.

\section{Evaluation Protocol}\label{app:evaluation_protocol}
\subsection{MetaTool}

In MetaTool evaluation, the agent receives (1) a user query, (2) a list of candidate tools with their names and descriptions, and (3) few-shot examples. The agent must select the appropriate tool name(s) from the provided list. For Similar, Scenario, and Multi-tool subtasks, the agent selects one or more ground-truth tools; for Reliability, the agent must output ``None'' when no suitable tool exists. Models must reason about and distinguish between all available tools in the candidate list, regardless of their names. The prompt is shown in Figure~\ref{fig:metatool-eval-prompt}.

\paragraph{Scoring Methodology.}  We revised the original MetaTool scoring methodology, which relied heavily on manual analysis and did not adequately handle ambiguous cases where models produced partially correct outputs. For single-tool tasks (Similar and Scenario), when exactly one
tool match is found, we verify that it corresponds to the
ground-truth label. If the keyword ``None'' also appears, the
response is marked correct only when the matched tool appears
before ``None'', as the MetaTool prompt requires the selected
tool to be stated first. For cases with two or more matches, we replaced manual analysis with automated evaluation using GPT-4.1-mini, which assesses whether the model's output selected only the ground-truth tools. For Reliability tasks, we added logic to verify correct ``None''
predictions: a response is marked correct if no matches are
found and ``None'' is present, or if matches exist but ``None''
appears before them. This distinguishes appropriate rejections
from cases where models incorrectly included tool names
alongside ``None''. For multi-tool tasks, responses with exactly one match are now explicitly marked incorrect, as these tasks require selecting multiple tools. Cases with more than two matches are evaluated using GPT-4.1-mini. These refinements enable consistent and scalable automated
scoring across all instances, which was difficult to achieve
with the original manual methodology.

\subsection{RoTBench}

In RoTBench evaluation, the agent receives (1) a system message describing the task format, (2) a list of available tools with complete schemas including names, descriptions, and parameter specifications, and (3) a user query. The agent must provide both tool selection and parameter identification in a structured format. The prompt is shown in Figure~\ref{fig:rotbench-eval-prompt}.

\paragraph{Evaluation Metrics.} Tool Selection accuracy is measured by whether the model selects the correct tool name from the available list. Parameter Identification accuracy is computed conditionally: only cases where the model correctly selected the tool are evaluated for whether the model also identified the correct set of required parameters. This two-stage evaluation isolates the impact of schema alignment on each capability, while preserving the multi-candidate selection challenge where models must reason across multiple tool options.

\subsection{API-Bank}\label{app:api_bank}

API-Bank~\citep{li-etal-2023-api} is a comprehensive benchmark for evaluating tool-augmented language models across realistic API usage scenarios. The benchmark implements 73 runnable APIs spanning diverse domains and evaluates models on 314 manually annotated dialogues containing 753 API calls.

API-Bank assesses two fundamental capabilities: (1) \textbf{Call}, where models must correctly invoke APIs with appropriate parameters when APIs are provided in the context, and (2) \textbf{Retrieval+Call}, where models must first retrieve relevant APIs from a large pool using an API Search function, then call the selected APIs. Both tasks require exact matching of tool names and parameters with ground-truth annotations for correctness. We evaluate on both capabilities to assess whether PA-Tool improves tool selection when schemas are either explicitly provided (Call) or must be retrieved (Retrieval+Call). The prompt is shown in Figure~\ref{fig:apibank-eval-prompt}.

\subsection{$\tau$-Bench}\label{app:tau_bench}

$\tau$-Bench~\citep{yao2025taubench} evaluates agents on realistic, multi-turn task-oriented dialogues where they must interact with simulated users while following domain-specific policies. Unlike traditional benchmarks that provide all information upfront, $\tau$-Bench requires agents to incrementally gather information through conversation, consult policy guidelines, and execute appropriate API calls to reach a target database state.

We evaluate on the Retail domain, which contains realistic customer service scenarios. Success requires: (1) correctly invoking APIs with write capabilities (e.g., \texttt{cancel\_order}, \texttt{update\_shipping}) to modify the database state, (2) adhering to domain-specific policies, and (3) managing multi-turn interactions to gather necessary information from users. The benchmark measures end-to-end task completion by comparing the final database state with the ground-truth state, directly assessing whether improved tool selection translates to successful task completion. We fix the user simulator to Llama3.1-8B-Instruct and use temperature 0 for deterministic evaluation, reporting the mean pass rate over 5 independent runs (N=5) to account for stochasticity in user simulation. The prompt is shown in Figure~\ref{fig:tau-bench-eval-prompt}.

\subsection{PA-Tool Integration}

Within these evaluation protocols, PA-Tool operates by replacing original tool names and parameter names in the provided schemas with pretraining-aligned alternatives. Critically, all other components remain identical: tool descriptions, parameter descriptions, example demonstrations, and task instructions are unchanged. This ensures that PA-Tool's improvements stem solely from schema alignment, not from modifications to semantic information or task structure. For instance, if the original schema contains \texttt{get\_weather\_info} but PA-Tool determines that \texttt{get\_weather\_forecast} better aligns with the model's pretrained knowledge, only the tool name in the tool schema is replaced. The tool's description, available parameters, and all contextual information remain identical across Base and PA-Tool conditions, ensuring fair and comparable evaluation.

\begin{table*}[ht!]
\centering
\resizebox{\textwidth}{!}{
\begin{tabular}{ll|cccc|cccc}
\toprule
\multirow{3}{*}{\textbf{Model}} & \multirow{3}{*}{\textbf{Method}} & \multicolumn{4}{c|}{\textbf{MetaTool}} & \multicolumn{4}{c}{\textbf{RoTBench}} \\
\cmidrule(lr){3-6} \cmidrule(lr){7-10}
& & \multicolumn{4}{c|}{Tool Selection} & \multicolumn{2}{c}{Single-turn} & \multicolumn{2}{c}{Multi-turn} \\
\cmidrule(lr){3-6} \cmidrule(lr){7-8} \cmidrule(lr){9-10}
& & Similar & Scenario & Reliability & Multi-tool & Tool Sel. & Param Iden. & Tool Sel. & Param Iden. \\
\midrule
\rowcolor{gray!15}
\multicolumn{10}{c}{\textit{Small Language Models}} \\
\midrule
\multirow{2}{*}{Ministral-8B} 
& Base & 65.3 & 81.2 & 42.5 & \textbf{93.1} & 66.7 & 28.6 & \textbf{51.4} & 38.6 \\
& PA-Tool & \textbf{70.5} & \textbf{82.6} & \textbf{44.7} & 92.4 & \textbf{75.2} & \textbf{31.4} & \textbf{51.4} & \textbf{41.4} \\
\midrule
\multirow{2}{*}{GPT-4.1-nano}
& Base & 56.8 & 53.1 & 99.0 & \textbf{81.5} & 71.4 & 34.3 & \textbf{51.4} & \textbf{44.3} \\
& PA-Tool & \textbf{60.6} & \textbf{54.0} & \textbf{99.5} & 77.1 & \textbf{72.4} & \textbf{44.8} & 47.1 & 42.9 \\
\midrule
\multirow{2}{*}{Gemini-2.5-Flash-Lite} 
& Base & 79.3 & 86.5 & 65.9 & \textbf{77.7} & 72.4 & 63.8 & \textbf{65.7} & \textbf{58.6} \\
& PA-Tool & \textbf{80.8} & \textbf{87.1} & \textbf{70.5} & 72.0 & \textbf{73.3} & \textbf{65.7} & 64.3 & \textbf{58.6} \\
\midrule
\rowcolor{gray!15}
\multicolumn{10}{c}{\textit{Large Language Models}} \\
\midrule
\multirow{2}{*}{Llama3.3-70B}
& Base & \textbf{80.4} & \textbf{85.2} & 63.7 & \textbf{82.9} & \textbf{75.2} & \textbf{27.6} & 60.0 & 47.1 \\
& PA-Tool & 79.9 & \textbf{85.2} & \textbf{66.2} & 81.3 & 74.3 & 26.7 & \textbf{64.3} & \textbf{48.6} \\
\midrule
\multirow{2}{*}{GPT-4.1-mini}
& Base & 79.6 & 84.3 & 76.3 & 72.2 & 79.1 & 58.1 & \textbf{71.4} & \textbf{61.4} \\
& PA-Tool & \textbf{80.5} & \textbf{84.5} & \textbf{80.0} & \textbf{84.3} & \textbf{82.9} & \textbf{59.1} & 70.0 & \textbf{61.4} \\
\midrule
\multirow{2}{*}{Gemini-2.5-Flash} 
& Base & 70.0 & 79.8 & 89.2 & 77.3 & 82.9 & 56.2 & 60.0 & 54.3 \\
& PA-Tool & \textbf{72.6} & \textbf{82.7} & \textbf{92.4} & \textbf{78.5} & \textbf{84.8} & \textbf{62.9} & \textbf{72.9} & \textbf{64.3} \\
\midrule
\rowcolor{gray!15}
\multicolumn{10}{c}{\textit{Reasoning Models (thinking mode)}} \\
\midrule
\multirow{2}{*}{Qwen3-1.7B}
& Base & 66.2 & \textbf{73.2} & 93.2 & \textbf{76.1} & 58.1 & 32.4 & 38.6 & \textbf{38.6} \\
& PA-Tool & \textbf{68.7} & 69.8 & \textbf{96.6} & 73.2 & \textbf{62.9} & \textbf{42.9} & \textbf{45.7} & \textbf{38.6} \\
\midrule
\multirow{2}{*}{Qwen3-4B}
& Base & 72.0 & \textbf{81.9} & 84.2 & \textbf{88.1} & 65.7 & \textbf{30.5} & 57.1 & 52.9 \\
& PA-Tool & \textbf{73.5} & 81.2 & \textbf{91.2} & 86.5 & \textbf{69.5} & 27.6 & \textbf{65.7} & \textbf{61.4} \\
\bottomrule
\end{tabular}
}
\caption{Performance comparison on MetaTool and RoTBench with additional models, grouped by model type. \textbf{Bold} indicates the better result between Base and PA-Tool per row group.}
\label{tab:additional_models}
\end{table*}

\section{Generalization to Diverse Models Details}\label{app:generalization_to_diverse_model}
To assess whether PA-Tool generalizes across different model families and scales, we conduct additional experiments on MetaTool and RoTBench using eight models beyond the Qwen2.5 and Llama families examined in our main results, bringing the total to 12 models across 5 families. We group these models into three categories. (1) \textbf{SLMs}: Ministral-8B, GPT-4.1-nano, and Gemini-2.5-Flash-Lite; (2) \textbf{LLMs}: Llama3.3-70B, GPT-4.1-mini, and Gemini-2.5-Flash, to test whether PA-Tool remains effective at larger scales; and (3) \textbf{reasoning models}: Qwen3-1.7B and Qwen3-4B in thinking mode, to examine whether inference-time chain-of-thought reasoning can independently resolve schema misalignment.

\paragraph{Results on SLMs.} Consistent with our main findings, PA-Tool improves performance over the Base in most settings. Gains are most pronounced for Ministral-8B, with a +8.5\% improvement on RoTBench single-turn tool selection (66.7\%→75.2\%) and +5.2\% on MetaTool Similar (65.3\%→70.5\%). For GPT-4.1-nano and Gemini-2.5-Flash-Lite, improvements are more task-dependent, with notable gains on MetaTool Reliability (+4.6\% for Gemini) and RoTBench parameter identification (+10.5\% for GPT-4.1-nano), though gains on other subtasks are smaller or absent.

\paragraph{Results on LLMs.} PA-Tool also improves larger models, though average gains are smaller than for SLMs, consistent with reduced prevalence of schema misalignment at larger scales. Gains remain substantial where misalignment compounds, such as multi-tool composition, where GPT-4.1-mini improves by +12.1\% on MetaTool Multi-tool (72.2\%→84.3\%). Llama3.3-70B shows the most modest improvements, suggesting that some larger open-source models may already handle misaligned schemas reasonably well.

\paragraph{Results on Reasoning Models.} Despite strong baselines from chain-of-thought reasoning (e.g., Qwen3-1.7B Reliability: 93.2\%), PA-Tool yields gains of up to +10.5\% on RoTBench single-turn parameter identification and +7.1\% on multi-turn tool selection for Qwen3-1.7B. Slight decreases on near-saturated subtasks (e.g., Qwen3-4B Multi-tool: 88.1\%→86.5\%) mirror patterns observed for larger non-reasoning models, where baselines are already strong. These results confirm that reasoning and schema alignment address complementary failure modes rather than competing solutions.

Together, these results demonstrate that PA-Tool generalizes across diverse model families, architectures, and scales.

%=============================================

\section{Integration with Supervised Fine-tuning Details}\label{app:sft_setup}
\paragraph{Setup.} As MetaTool does not include a predefined training split, we construct train\slash validation \slash test sets by randomly sampling 60\%/20\%/20\% of all task instances. Since the dataset provides only ground-truth labels without reasoning traces, we use GPT-4.1-mini to generate reasoning trajectories for each training instance (prompt in Figure~\ref{fig:reasoning-generation-prompt}), which we use to train the SFT1 baseline. SFT2 doubles this training data to test whether PA-Tool remains effective under larger training sets. We fine-tune Llama3.1-8B with the hyperparameters in Table~\ref{tab:training_setup}, and additionally fine-tune Qwen2.5-3B and Qwen2.5-7B under the SFT1 configuration to verify generalization across model families.

\paragraph{Results.} Table~\ref{tab:sft_qwen} presents the performance on Qwen2.5 models. The results confirm the key findings from Llama3.1-8B (Table~\ref{tab:sft}): SFT strengthens tool-use reasoning through training data, while PA-Tool resolves naming-level misalignment that persists even after fine-tuning. SFT + PA-Tool achieves the best results on most subtasks (e.g., Qwen2.5-3B Reliability: 89.5\%, Multi-tool: 77.8\%), generalizing the complementary gains observed on Llama3.1-8B to the Qwen2.5 family.

\begin{table}[t]
    \centering
    \resizebox{\columnwidth}{!}{
    \begin{tabular}{llcccc}
        \hline
        \textbf{Model} & \textbf{Config} & \textbf{Similar} & \textbf{Scenario} & \textbf{Reliab.} & \textbf{Multi.} \\
        \hline
        \multirow{4}{*}{Qwen2.5-3B}
         & Base          & 50.3 & 61.2 & 84.9 & 63.6 \\
         & PA-Tool       & 50.8 & \textbf{64.4} & 83.9 & 68.7 \\
         & SFT           & 50.8 & 62.3 & 85.4 & 64.6 \\
         & SFT + PA-Tool & \textbf{51.8} & 59.8 & \textbf{89.5} & \textbf{77.8} \\
        \hline
        \multirow{4}{*}{Qwen2.5-7B}
         & Base          & \textbf{61.8} & \textbf{80.3} & 80.9 & 79.8 \\
         & PA-Tool       & 60.8 & 78.6 & 84.9 & 85.9 \\
         & SFT           & 60.8 & 80.2 & 81.9 & 82.8 \\
         & SFT + PA-Tool & \textbf{61.8} & 78.6 & \textbf{86.9} & \textbf{86.9} \\
        \hline
    \end{tabular}
    }
    \caption{SFT comparison on Qwen2.5 models.}
    \label{tab:sft_qwen}
\end{table}

\begin{table}[h]
\centering
\small
\begin{tabular}{ll}
\toprule
\textbf{Hyperparameter} & \textbf{Value} \\
\midrule
LoRA Rank & 32 \\
LoRA Alpha & 64 \\
LoRA Dropout & 0.05 \\
Learning Rate & 5e-5 \\
Batch Size & 16 \\
Epochs & 5 \\
GPU & NVIDIA A100 80GB PCIe \\
\bottomrule
\end{tabular}
\caption{Training hyperparameters for SFT.}
\label{tab:training_setup}
\end{table}

%=============================================

\section{Error Analysis Details}
\label{app:error-analysis}
\paragraph{Error Taxonomy.} We categorize tool selection errors into three types based on their underlying failure modes:

\textbf{(1) Schema Misalignment Error}: the model generates a plausible but non-existent tool name, following its pretrained naming conventions rather than the provided schema.

\textbf{(2) Functional Confusion Error}: the model selects an existing tool whose functionality is similar to the correct one, indicating it understands the query but confuses related tools. For instance, when asked to send an email notification, the model might select \texttt{send\_sms} instead of \texttt{send\_email}.

\textbf{(3) Context Understanding Error}: the model selects a functionally unrelated tool, indicating a failure to comprehend the query's intent. For example, when asked to delete a user account, the model might select \texttt{create\_user} or \texttt{list\_products}.

\paragraph{Experimental Setup.} We use GPT-4.1-mini \cite{openai_gpt41mini_2024} as an error analyzer to classify errors from Llama3.1-8B in both Base and PA-Tool configurations. We analyze MetaTool's three tool selection subtasks---Similar, Scenario, and Multi-tool---excluding Reliability as it specifically tests the ``no suitable tool'' scenario rather than tool selection errors. The prompt is provided in Figure~\ref{fig:error-classification-prompt}.

%=============================================

\section{Computational Time}\label{app:computational_time}

\begin{table}[h]
\centering{
\small
\begin{tabular}{lcc}
\toprule
\textbf{Model} & \textbf{MetaTool} & \textbf{RoTBench} \\
\midrule
Qwen2.5-3B & 8.0 sec & 13.9 sec \\
Qwen2.5-7B & 11.2 sec & 15.0 sec \\
Llama3.2-3B & 8.2 sec & 12.4 sec \\
Llama3.1-8B & 11.6 sec & 15.9 sec \\
\bottomrule
\end{tabular}
}
\caption{Schema generation time for PA-Tool.}
\label{tab:computational_time}
\end{table}

Table~\ref{tab:computational_time} shows the wall-clock time required for one-time schema generation on both MetaTool (199 tools) and RoTBench (568 tools). Schema generation is a one-time preprocessing cost, requiring 8--16 seconds depending on model size and number of tools. Once generated, the pretraining-aligned schema can be reused indefinitely without additional overhead during inference. All experiments were conducted on Intel Xeon Gold 6230 CPU @ 2.10GHz (38 cores), NVIDIA A100 80GB PCIe, 450GB RAM, with Python 3.11.14, vLLM 0.10.2, and CUDA 12.8.

%=============================================

%=============================================

\begin{table*}[ht!]
\centering
\resizebox{\textwidth}{!}{
\begin{tabular}{ll|cccc|cccc}
\toprule
\multirow{3}{*}{\textbf{Model}} & \multirow{3}{*}{\textbf{Method}} & \multicolumn{4}{c|}{\textbf{MetaTool}} & \multicolumn{4}{c}{\textbf{RoTBench}} \\
\cmidrule(lr){3-6} \cmidrule(lr){7-10}
& & \multicolumn{4}{c|}{Tool Selection} & \multicolumn{2}{c}{Single-turn} & \multicolumn{2}{c}{Multi-turn} \\
\cmidrule(lr){3-6} \cmidrule(lr){7-8} \cmidrule(lr){9-10}
& & Similar & Scenario & Reliability & Multi-tool & Tool Sel. & Param Iden. & Tool Sel. & Param Iden. \\
\midrule

\multirow{6}{*}{Qwen2.5-3B} & Base & 48.7 & 55.3 & 83.6 & \textbf{75.1} & 12.4& 7.6& 10.0& 10.0\\
 & BM25 & 48.7 & 55.3 & 83.6 & \textbf{75.1} & 14.3& 8.6& 12.9& 11.4\\
 & ToolLLM & 48.7 & 55.3 & 83.7 & \textbf{75.1} & 4.8 & 1.9 & 12.9& 11.4\\
 & PA-Tool & 50.0 & 58.8 & \textbf{86.2} & 72.6 & 18.1& 10.5& 15.7& 14.3\\
 & PA-Tool + BM25 & \textbf{50.3} & 60.2 & \textbf{86.2} & 72.8 & \textbf{21.9}& \textbf{13.3}& 20.0& 17.1\\
 & PA-Tool + ToolLLM & \textbf{50.3} & \textbf{60.3} & \textbf{86.2} & 73.4 & 20.9& \textbf{13.3}& \textbf{21.4}& \textbf{20.0}\\
\midrule

\multirow{6}{*}{Qwen2.5-7B} & Base & 59.6 & 74.4 & 78.3 & 78.3 & 49.5 & 20.0 & 21.4& 21.4\\
 & BM25 & 59.8 & 74.6 & 78.3 & 78.3 & 49.5 & 20.9 & 21.4& 21.4\\
 & ToolLLM & 59.8 & 74.6 & 78.3 & 78.3 & 49.5 & 20.9 & 21.4& 21.4\\
 & PA-Tool & \textbf{64.1} & 78.4 & \textbf{88.2} & 84.9 & 55.2 & \textbf{21.9} & \textbf{27.1}& \textbf{22.9}\\
 & PA-Tool + BM25 & \textbf{64.1} & \textbf{78.5} & \textbf{88.2} & 84.9 & \textbf{56.2} & \textbf{21.9} & \textbf{27.1}& \textbf{22.9}\\
 & PA-Tool + ToolLLM & \textbf{64.1} & \textbf{78.5} & \textbf{88.2} & \textbf{85.5} & 55.2 & \textbf{21.9} & \textbf{27.1}& \textbf{22.9}\\
\midrule

\multirow{6}{*}{Llama3.2-3B} & Base & 55.0 & 58.6 & 43.6 & 79.1 & 56.2 & 20.0 & 32.9& 27.1\\
 & BM25 & 55.7 & 58.7 & 43.6 & 79.1 & 56.2 & 20.9& 32.9& 27.1\\
 & ToolLLM & 55.4 & 58.8 & 43.6 & 79.1 & 56.2 & \textbf{22.9} & 32.9& 27.1\\
 & PA-Tool & \textbf{65.7} & \textbf{67.7} & \textbf{60.6} & \textbf{80.5} & 62.9 & 21.9 & \textbf{34.3}& \textbf{28.6}\\
 & PA-Tool + BM25 & \textbf{65.7} & \textbf{67.7} & \textbf{60.6} & \textbf{80.5} & \textbf{63.8} & 21.9& \textbf{34.3}& \textbf{28.6}\\
 & PA-Tool + ToolLLM & \textbf{65.7} & \textbf{67.7} & \textbf{60.6} & \textbf{80.5} & 62.9 & 21.9 & \textbf{34.3}& \textbf{28.6}\\
\midrule

\multirow{6}{*}{Llama3.1-8B} & Base & 61.5 & 73.9 & 53.5 & 78.7 & 58.1 & 17.1 & 42.8& 34.3\\
 & BM25 & 61.8 & 74.0 & 53.5 & 78.7 & 58.1 & 17.1& 42.8& 34.3\\
 & ToolLLM & 61.6 & 74.0 & 53.5 & 78.7 & 58.1 & 17.1& 42.8& 34.3\\
 & PA-Tool & \textbf{70.4} & \textbf{79.9} & \textbf{66.0} & \textbf{88.3} & \textbf{68.6} & \textbf{18.1} & \textbf{48.6}& 35.7\\
 & PA-Tool + BM25 & \textbf{70.4} & \textbf{79.9} & \textbf{66.0} & \textbf{88.3} & \textbf{68.6} & \textbf{18.1} & \textbf{48.6}& \textbf{37.1}\\
 & PA-Tool + ToolLLM & \textbf{70.4} & \textbf{79.9} & \textbf{66.0} & \textbf{88.3} & \textbf{68.6} & \textbf{18.1}& \textbf{48.6}& \textbf{37.1}\\
\bottomrule
\end{tabular}
}
\caption{Performance of retrieval-based post-hoc correction methods (BM25, ToolLLM) on MetaTool and RoTBench.}
\label{tab:postprocessing}
\end{table*}
\section{Integration with Training-free Methods Details}\label{app:comparisonwithothermodels}
\subsection{Retrieval-based Correction}
Retrieval-based methods recover from schema misalignment by mapping invalid model outputs to valid tool names post-hoc. When a model prediction is not found in the candidate list (case-insensitive check), we retrieve the nearest valid tool; predictions already within the candidate list are left unchanged, even if semantically incorrect. This distinction is important: retrieval can only recover schema-violating outputs, not wrong selections among valid tools.

\paragraph{BM25.} We retrieve the most similar valid tool using BM25 scoring~\citep{bm25} based on lexical overlap.

\paragraph{ToolLLM Embedding.} We employ ToolLLM's tool-specialized embedding model~\citep{qin2024toolllm} and retrieve the nearest valid tool in embedding space.

\paragraph{Results.} 
As shown in Table~\ref{tab:postprocessing}, retrieval-based correction yields limited improvements on its own. On MetaTool, BM25 and ToolLLM produce marginal or no gains in most subtasks, with accuracy improvements rarely exceeding 1\%. On RoTBench, gains are slightly larger but remain inconsistent across models and settings. When combined with PA-Tool, however, retrieval methods yield further improvements on top of PA-Tool's gains, indicating that schema alignment and post-hoc retrieval address different aspects of tool-use and can be combined for additional benefit.

\begin{table*}[!t]
\centering
\resizebox{\textwidth}{!}{
\begin{tabular}{ll|cccc|cccc}
\toprule
\multirow{3}{*}{\textbf{Model}} & \multirow{3}{*}{\textbf{Method}} & \multicolumn{4}{c|}{\textbf{MetaTool}} & \multicolumn{4}{c}{\textbf{RoTBench}} \\
\cmidrule(lr){3-6} \cmidrule(lr){7-10}
& & \multicolumn{4}{c|}{Tool Selection} & \multicolumn{2}{c}{Single-turn} & \multicolumn{2}{c}{Multi-turn} \\
\cmidrule(lr){3-6} \cmidrule(lr){7-8} \cmidrule(lr){9-10}
& & Similar & Scenario & Reliability & Multi-tool & Tool Sel. & Param Iden. & Tool Sel. & Param Iden. \\
\midrule
\multirow{4}{*}{Qwen2.5-3B}
& Base & 48.7 & 55.3 & 83.6 & 75.1 & 12.4 & 7.6 & 10.0 & 10.0 \\
& PA-Tool & 50.0 & 58.8 & 86.2 & 72.6 & 18.1 & 10.5 & 15.7 & 14.3 \\

& Constrained & 65.9 & \textbf{80.3} & 60.9 & \textbf{77.3} & \textbf{65.7} & 16.2 & \textbf{42.9}& \textbf{32.9}\\
& Constrained-PA & \textbf{66.1} & 77.9 & \textbf{74.5} & 76.5 & 61.9 & \textbf{26.7} & 37.1& 30.0\\
\midrule
\multirow{4}{*}{Qwen2.5-7B} 
& Base & 59.6 & 74.4 & 78.3 & 78.3 & 49.5 & 20.0 & 21.4 & 21.4 \\
& PA-Tool & 64.1 & 78.4 & \textbf{88.2} & \textbf{84.9} & 55.2 & 21.9 & 27.1 & 22.9 \\

& Constrained & 64.2 & \textbf{81.0} & 78.4 & 83.3 & \textbf{78.1} & \textbf{31.4} & \textbf{71.4}& 51.4\\
& Constrained-PA & \textbf{67.9} & 79.3 & 84.0 & 84.1 & 76.2 & \textbf{31.4} & 67.1& \textbf{54.3}\\
\midrule
\multirow{4}{*}{Llama3.2-3B}
& Base & 55.0 & 58.6 & 43.6 & 79.1 & 56.2 & 20.0 & 32.9 & 27.1 \\
& PA-Tool & \textbf{65.7} & 67.7 & 60.6 & 80.5 & 62.9 & \textbf{21.9} & 34.3 & 28.6 \\

& Constrained & 55.1 & \textbf{70.2} & \textbf{72.3} & \textbf{89.5} & 62.9 & 20.0 & 41.4& 31.4\\
& Constrained-PA & 59.2 & 69.8 & 65.5 & 86.9 & \textbf{68.6} & \textbf{21.9} & \textbf{44.3}& \textbf{40.0}\\
\midrule
\multirow{4}{*}{Llama3.1-8B} 
& Base & 61.5 & 73.9 & 53.5 & 78.7 & 58.1 & 17.1 & 42.8 & 34.3 \\
& PA-Tool & \textbf{70.4} & 79.9 & 66.0 & \textbf{88.3} & \textbf{68.6} & 18.1 & \textbf{48.6} & \textbf{35.7} \\

& Constrained & 66.3 & \textbf{81.7} & 66.2 & 74.3 & 64.8 & 16.2 & 44.3& \textbf{35.7}\\
& Constrained-PA & 69.8 & 81.2 & \textbf{66.3} & 79.9 & 63.8 & \textbf{20.0} & 45.7& \textbf{35.7}\\
\bottomrule
\end{tabular}
}
\caption{Performance of \textbf{Constrained} generation using JSON Schema enum on MetaTool and RoTBench}
\label{tab:constrained}
\end{table*}

\subsection{Constrained Generation}

Constrained generation methods enforce valid outputs during the decoding process, preventing schema-violating generations.

\paragraph{JSON Schema Constraints.} We implement constrained decoding using JSON Schema with enum restrictions, which masks logits corresponding to tokens that would lead to invalid tool names during generation. For MetaTool, the schema restricts outputs to \texttt{\{"selected\_tool": <enum>\}}, while for RoTBench it enforces \texttt{\{"action": <enum>, "action\_input": <object>\}}, where the enum contains all valid tool names plus special tokens (e.g., \texttt{None}, \texttt{finish}).

\paragraph{Results.}
As shown in Table~\ref{tab:constrained}, constrained generation substantially improves RoTBench performance by eliminating format errors (e.g., Qwen2.5-7B single-turn tool selection: 49.5\%→78.1\%). On MetaTool, however, results are more mixed: constrained generation sometimes underperforms PA-Tool alone, and can even fall below Base on Reliability (e.g., Qwen2.5-3B: 83.6\%→60.9\%), likely because enum restrictions bias the model toward emitting a tool even when none is appropriate. Combining constrained generation with PA-Tool (Constrained-PA) mitigates this: the approach retains format guarantees while leveraging PA-Tool's schema alignment, yielding the best parameter identification results on most models (e.g., Qwen2.5-3B single-turn: 16.2\%→26.7\%). This suggests that PA-Tool's aligned representations help the model select correctly even when the output space is explicitly constrained.

\begin{table*}[ht!]
\centering
\resizebox{\textwidth}{!}{
\begin{tabular}{ll|cccc|cccc}
\toprule
\multirow{3}{*}{\textbf{Model}} & \multirow{3}{*}{\textbf{Method}} & \multicolumn{4}{c|}{\textbf{MetaTool}} & \multicolumn{4}{c}{\textbf{RoTBench}} \\
\cmidrule(lr){3-6} \cmidrule(lr){7-10}
& & \multicolumn{4}{c|}{Tool Selection} & \multicolumn{2}{c}{Single-turn} & \multicolumn{2}{c}{Multi-turn} \\
\cmidrule(lr){3-6} \cmidrule(lr){7-8} \cmidrule(lr){9-10}
& & Similar & Scenario & Reliability & Multi-tool & Tool Sel. & Param Iden. & Tool Sel. & Param Iden. \\
\midrule

\multirow{4}{*}{Qwen2.5-3B}
& Base & 48.7 & 55.3 & 83.6 & \textbf{75.1} & 12.4 & 7.6 & 10.0 & 10.0 \\
& PA-Tool & 50.0 & 58.8 & 86.2 & 72.6 & 18.1 & 10.5 & 15.7 & \textbf{14.3} \\

& EasyTool & \textbf{52.3} & \textbf{64.3} & 91.5 & 70.6 & 19.1 & 12.4 & 14.3 & 11.4 \\
& EasyTool+PA-Tool & 48.0 & 47.7 & \textbf{93.4} & 46.1 & \textbf{22.9} & \textbf{19.1} & \textbf{17.1} & \textbf{14.3} \\
\midrule
\multirow{4}{*}{Qwen2.5-7B} 
& Base & 59.6 & 74.4 & 78.3 & 78.3 & 49.5 & 20.0 & 21.4 & 21.4 \\
& PA-Tool & 64.1 & 78.4 & 88.2 & 84.9 & 55.2 & 21.9 & \textbf{27.1} & 22.9 \\
& EasyTool & \textbf{65.1} & \textbf{78.9} & 83.0 & \textbf{85.1} & 50.5 & \textbf{26.7} & 25.7 & 24.3 \\
& EasyTool+PA-Tool & 64.5 & 73.5 & \textbf{90.5} & 79.3 & \textbf{52.4} & \textbf{26.7} & \textbf{27.1} & \textbf{25.7} \\
\midrule
\multirow{4}{*}{Llama3.2-3B}
& Base & 55.0 & 58.6 & 43.6 & 79.1 & 56.2 & 20.0 & 32.9 & 27.1 \\
& PA-Tool & \textbf{65.7} & \textbf{67.7} & 60.6 & \textbf{80.5} & 62.9 & 21.9 & 34.3 & 28.6 \\
& EasyTool & 45.3 & 65.3 & 64.1 & 79.5 & 56.2 & 20.9 & 35.7 & 28.6 \\
& EasyTool+PA-Tool & 42.0 & 58.3 & \textbf{80.5} & 66.6 & \textbf{66.7} & \textbf{23.8} & \textbf{38.6} & \textbf{30.0} \\
\midrule
\multirow{4}{*}{Llama3.1-8B} 
& Base & 61.5 & 73.9 & 53.5 & 78.7 & 58.1 & 17.1 &  42.8 & 34.3 \\
& PA-Tool & \textbf{70.4} & \textbf{79.9} & 66.0 & \textbf{88.3} & \textbf{68.6} & 18.1 & \textbf{48.6} & 35.7 \\
& EasyTool & 61.1 & 77.2 & \textbf{76.9} & 72.2 & 59.0 & 14.3 & 40.0 & 31.4 \\
& EasyTool+PA-Tool & 66.2 & 77.9 & 73.8 & 68.6 & 64.8 & \textbf{21.9} & 41.4 & \textbf{37.1} \\
\bottomrule
\end{tabular}
}
\caption{Performance of description enhancement (\textbf{EasyTool}) on MetaTool and RoTBench}
\label{tab:easytool}
\end{table*}
\subsection{Description Enhancement}

Description enhancement methods improve tool understanding by rewriting documentation for clarity, without modifying the tool naming schema.

\paragraph{EasyTool.} EasyTool~\citep{yuan-etal-2025-easytool} creates unified tool instructions by restructuring verbose API documentation into clear, consistent formats. We adapt this methodology with two modifications. First, to ensure fair comparison without introducing external model capabilities, we use the \textit{target model itself} rather than GPT-4 to generate descriptions. Second, we simplify the output to concise 1--2 sentence descriptions with action-oriented verbs (e.g., ``Retrieves'', ``Calculates''). The full prompt template follows EasyTool's principles of removing redundancy and improving clarity (Figure~\ref{fig:easytool-prompt}).

\paragraph{Results.} 
As shown in Table~\ref{tab:easytool}, PA-Tool and EasyTool show complementary strengths across models. On Llama models, PA-Tool substantially outperforms EasyTool (e.g., Llama3.2-3B Similar: 65.7\% vs.\ 45.3\%; Llama3.1-8B Multi-tool: 88.3\% vs.\ 72.2\%), suggesting that naming alignment drives larger gains than description enhancement for these models. On Qwen models, EasyTool is competitive and sometimes stronger, particularly for Qwen2.5-3B on MetaTool subtasks (e.g., Scenario: 64.3\% vs.\ 58.8\%), indicating that different models benefit from different axes of schema improvement. Importantly, the two approaches are compatible: combining EasyTool with PA-Tool achieves the best Reliability scores on three of four models (e.g., Qwen2.5-3B: 93.4\%), and yields the strongest overall parameter identification on RoTBench (e.g., Qwen2.5-3B single-turn: 19.1\%), showing that name alignment and description clarity address orthogonal aspects of schema quality.

\begin{table}[t]
    \centering
    \resizebox{\columnwidth}{!}{
    \begin{tabular}{llcccc}
        \toprule
        \textbf{Target} & \textbf{Schema Source} & \textbf{Similar} & \textbf{Scenario} & \textbf{Reliab.} & \textbf{Multi.} \\
        \midrule
        \multirow{5}{*}{Llama3.2-3B}
         & Base (no PA-Tool)   & 55.0 & 58.6 & 43.6 & 79.1 \\
         & Qwen2.5-7B          & 61.6 & 59.3 & \textbf{65.5} & 80.7 \\
         & Qwen2.5-3B          & 58.4 & 60.2 & 67.6 & 79.3 \\
         & Llama3.1-8B         & 63.3 & 65.9 & 67.5 & \textbf{82.7} \\
         & Self                & \textbf{65.7} & \textbf{67.7} & 60.6 & 80.5 \\
        \midrule
        \multirow{5}{*}{Llama3.1-8B}
         & Base (no PA-Tool)   & 61.5 & 73.9 & 53.5 & 78.7 \\
         & Qwen2.5-7B          & 69.6 & 77.4 & 54.1 & 83.1 \\
         & Qwen2.5-3B          & 68.9 & 78.2 & 54.7 & 82.5 \\
         & Llama3.2-3B         & 69.2 & 79.2 & 51.2 & 85.7 \\
         & Self                & \textbf{70.4} & \textbf{79.9} & \textbf{66.0} & \textbf{88.3} \\
        \bottomrule
    \end{tabular}
    }
    \caption{Cross-model schema transfer on MetaTool. \textit{Self} denotes the target model's own PA-Tool schema.}
    \label{tab:cross_model}
\end{table}

\section{Cross-Model Schema Transfer}\label{app:cross_model}

PA-Tool generates model-specific schemas, raising a practical
question: in multi-agent systems with heterogeneous SLMs, must
each model use its own schema? We investigate cross-model schema
transfer by evaluating target models with schemas generated by
different source models.

\paragraph{Results.} Table~\ref{tab:cross_model} presents
cross-model schema transfer results on MetaTool. We observe three
findings. First, cross-model schemas improve over
unaligned baselines, even across model families (e.g.,
Llama3.1-8B with Qwen2.5-7B schemas achieves 69.6\% on Similar,
+8.1\% over Base), suggesting partially overlapping naming
conventions due to shared training corpora. Second, self-generated
schemas are optimal in most settings, but cross-model gaps are task-dependent: on
Similar and Scenario, gaps are within 1--3\%, while Reliability
shows larger gaps (e.g., 54.1\% vs.\ 66.0\% for Llama3.1-8B),
indicating that tasks requiring precise schema understanding are
more sensitive to model-specific alignment. Third, larger models'
schemas can benefit smaller models within the same family;
Llama3.2-3B with Llama3.1-8B schemas achieves the highest
scores on Multi-tool among cross-model
configurations.

We recommend generating model-specific schemas given the
negligible one-time cost (Appendix~\ref{app:computational_time}). When this is infeasible, cross-model
schemas---particularly from a larger model in the same
family---provide a viable fallback that still outperforms
unaligned baselines.

%=============================================
\begin{table}[t]
\centering
\scriptsize
\setlength{\tabcolsep}{4pt}
\begin{tabular}{ll|cc|cc}
\toprule
\multirow{2}{*}{\textbf{Model}} & \multirow{2}{*}{\textbf{Level}} & \multicolumn{2}{c|}{\textbf{Single-turn}} & \multicolumn{2}{c}{\textbf{Multi-turn}} \\
\cmidrule(lr){3-4} \cmidrule(lr){5-6}
& & Tool Sel. & Param Id. & Tool Sel. & Param Id. \\
\midrule
\multirow{6}{*}{Qwen2.5-3B}
 & Clean   & 12.4 & 7.6  & 10.0 & 10.0 \\
 & Slight  & 13.3 & 6.2  & 7.9  & 6.4  \\
 & Medium  & 11.4 & 6.7  & 7.1  & 5.0  \\
 & Heavy   & \textbf{21.9} & 5.7  & 9.3  & 5.0  \\
 & Union   & 8.6  & 3.8  & 10.0 & 4.3  \\
 & PA-Tool & 18.1 & \textbf{10.5} & \textbf{15.7} & \textbf{14.3} \\
\midrule
\multirow{6}{*}{Qwen2.5-7B}
 & Clean   & 49.5 & 20.0 & 21.4 & 21.4 \\
 & Slight  & 33.3 & 12.9 & 20.0 & 18.6 \\
 & Medium  & 28.1 & 10.0 & 16.4 & 15.0 \\
 & Heavy   & 47.6 & 12.9 & \textbf{27.9} & 18.6 \\
 & Union   & 41.9 & 17.1 & 25.7 & 20.0 \\
 & PA-Tool & \textbf{55.2} & \textbf{21.9} & 27.1 & \textbf{22.9} \\
\midrule
\multirow{6}{*}{Llama3.2-3B}
 & Clean   & 56.2 & 20.0 & 32.9 & 27.1 \\
 & Slight  & 34.8 & 10.5 & 20.0 & 13.6 \\
 & Medium  & 31.9 & 10.5 & 17.1 & 10.7 \\
 & Heavy   & 47.6 & 11.9 & 23.6 & 12.9 \\
 & Union   & 50.5 & 16.2 & 21.4 & 14.3 \\
 & PA-Tool & \textbf{62.9} & \textbf{21.9} & \textbf{34.3} & \textbf{28.6} \\
\midrule
\multirow{6}{*}{Llama3.1-8B}
 & Clean   & 58.1 & 17.1 & 42.8 & 34.3 \\
 & Slight  & 37.1 & 8.1  & 27.9 & 22.1 \\
 & Medium  & 31.9 & 8.1  & 25.7 & 21.4 \\
 & Heavy   & 48.6 & 10.0 & 31.4 & 23.6 \\
 & Union   & 46.7 & 12.4 & 40.0 & 25.7 \\
 & PA-Tool & \textbf{68.6} & \textbf{18.1} & \textbf{48.6} & \textbf{35.7} \\
\bottomrule
\end{tabular}
\caption{Performance across RoTBench noise levels.}
\label{tab:noise_levels}
\vspace{-1em}
\end{table}

\section{Robustness Across Noise Levels}\label{app:noise_levels}
A central premise of PA-Tool is that SLMs rely heavily on tool \emph{names}, not just descriptions, when making tool-use decisions. To stress-test this premise and demonstrate PA-Tool's robustness, we evaluate all four main models across the five noise environments defined by RoTBench~\cite{ye-etal-2024-rotbench}: \textbf{Clean} (original schemas), \textbf{Slight} (character-level insertions, omissions, and substitutions), \textbf{Medium} (names replaced with reversed strings or random characters such as ``abc''), \textbf{Heavy} (names shuffled across tools, creating anti-semantic assignments), and \textbf{Union} (a random combination of the above perturbations to both tool names and parameters). PA-Tool generates schemas solely from descriptions, so its output is identical regardless of the original schema's noise level.

\paragraph{Results.}
Table~\ref{tab:noise_levels} reports single-turn and multi-turn results across all noise levels. Three patterns emerge, together motivating PA-Tool's role.

First, \textbf{tool names are primary selection cues}. For most models, even the mildest perturbation (Slight) causes large drops in single-turn tool selection despite descriptions remaining intact (e.g., Llama3.1-8B: $58.1\% \to 37.1\%$; Llama3.2-3B: $56.2\% \to 34.8\%$). 

Second, \textbf{not all names are equally useful: plausibility matters, but alignment matters more}. Medium noise, where names become semantically meaningless, produces the most severe drops (e.g., Llama3.1-8B: $58.1\% \to 31.9\%$), while Heavy noise shows partial recovery (e.g., $58.1\% \to 48.6\%$) because shuffled names still preserve real-world naming patterns even when attached to the wrong tools. Models thus benefit from \emph{plausible} names, but \emph{correctly aligned} names are needed for full performance---and human-designed Clean schemas are not guaranteed to provide them.

Third, \textbf{PA-Tool provides such alignment, surpassing the Clean baseline on nearly all settings}. Single-turn tool-selection gains reach $+5.7$ to $+10.5$ points on the three stronger models (e.g., Llama3.1-8B: $58.1\% \to 68.6\%$; Qwen2.5-7B: $49.5\% \to 55.2\%$; Llama3.2-3B: $56.2\% \to 62.9\%$), with consistent improvements on multi-turn settings and parameter identification. Since PA-Tool generates names from descriptions alone, it yields the same pretraining-aligned schema regardless of input quality, providing a stable performance floor for real-world deployments where tool schemas may be inconsistently named or poorly documented.

\section{AI Assistants in Research or Writing}
We used AI assistants to refine writing, proofread the text, and assist with coding experiments. However, all core ideas, experimental design, analysis, and scientific contributions are entirely the work of the authors.

\section{Potential Risks}
\label{app:risks}

We acknowledge two potential risks associated with deploying PA-Tool.
First, because PA-Tool explicitly aligns tool schemas with patterns
internalized during pretraining, it may preserve and reinforce biased,
culturally narrow, or English-centric naming conventions present in the
pretraining corpus; practitioners deploying PA-Tool on multilingual or
domain-specific tool ecosystems may need to audit the generated names
for cultural appropriateness and inclusivity. Second, although our human
evaluation (\S\ref{sec:human_eval}) shows that PA-Tool names are
generally rated as more understandable than the originals, a minority
of cases ($\sim$10\%) yield names that human developers find less clear, and a human-in-the-loop review of
the generated schema mapping may be warranted before deployment in
safety-critical applications.
 
\section{Prompt Templates}
\label{app:prompt}
\begin{figure}[h]
\centering
\begin{tcolorbox}[title=Candidate Name Generation Prompt, halign=left, boxrule=0.5pt]\footnotesize
Generate a \texttt{\{\{ component \}\}} name from the description below. \\
The \texttt{\{\{ component \}\}} will be used in a tool agent scenario.
\vspace{2ex}

Description: \\
\texttt{\{\{ description \}\}}
\vspace{2ex}

\textbf{If component == "tool":}
\vspace{1ex}

Example: \\
Description: A tool that manages files and directories on the system. \\
Output: \texttt{file\_manager}
\vspace{1ex}

Generate only the name without additional explanation.
\vspace{2ex}

\textbf{Elif component == "parameter":}
\vspace{1ex}

Example: \\
Context: \\
Tool: \texttt{file\_manager} - A tool for managing files and directories \\
Output: \texttt{file\_path}
\vspace{1ex}

\vspace{1ex}
Context:\\
Tool: \texttt{\{\{ tool\_name \}\}} - {\{\{ tool\_description \}\}}

Generate only the name without additional explanation.

\end{tcolorbox}
\caption{Candidate name generation prompt.}
\label{fig:component-generation-prompt}
\end{figure}

\begin{figure}[!h]\label{rotbench_inference}
\begin{tcolorbox}[title=RoTBench Inference Prompt, halign=left, boxrule=0.5pt]\footnotesize
\textbf{System:} You are an expert in using tools. At each step, analyze the state and decide the next action with a function call.
\vspace{1ex}

\textbf{Available tools:} \\
\texttt{[\{"name": "get\_translation\_nllb",} \\
\texttt{~~"description": "Translate text using NLLB model.",} \\
\texttt{~~"parameters": \{"input\_text": \{...\}, "tgt\_lang": \{...\},} \\
\texttt{~~~"src\_lang": \{...\}, ...\}\},} \\
\texttt{~\{"name": "get\_translation\_baidu",} \\
\texttt{~~"description": "Translate using BAIDU API.",} \\
\texttt{~~"parameters": \{"text": \{...\}, "tgt\_lang": \{...\}, ...\}\},}...] \\
\vspace{1ex}

\textbf{User:} ``What is the translation of `See you later' in Japanese?''
\vspace{1ex}

\textbf{Expected output:} \\
Thought: Need to translate English to Japanese \\
Action: get\_translation\_nllb \\
Action Input: \{"input\_text": "See you later", "tgt\_lang": "jpn\_Jpan", "src\_lang": "eng\_Latn"\}

\end{tcolorbox}
\caption{RoTBench inference prompt with system message, tool schemas, and structured output format.}
\label{fig:rotbench-eval-prompt}
\end{figure}

\begin{figure}[h]
\begin{tcolorbox}[title=MetaTool Inference Prompt, halign=left, boxrule=0.5pt]\footnotesize
You are a helpful AI assistant. Your current task is to choose the appropriate tool to solve the user's query based on their question. I will provide you with the user's question and information about the tools. If there is a tool in the list that is applicable to this query, please return the name of the tool (you can only choose one tool). If there isn't, please return `None.' Additionally, you will need to support your answer with a brief explanation.
\vspace{2ex}

\textbf{User's Query:} [User's Query Start] Planning a beach day next week, while considering the ideal weather conditions, availability of beach accessories, preferred beach location, potential activities to engage in like swimming, sunbathing, or playing beach games, and coordinating with friends or family members to join in the fun. [User's Query End]
\vspace{2ex}

\textbf{List of Tools with Names and Descriptions:} [List of Tools with Names and Descriptions Start]
\begin{enumerate}
    \item tool name: airqualityforeast, tool description: [`Planning something outdoors? Get the 2-day air quality forecast for any US zip code.']
    \item tool name: WeatherTool, tool description: Provide you with the latest weather information.
    \item tool name: AusSurfReport, tool description: [`Get today's surf report for any break throughout Australia!']
    \item[] ...
    \item[10.] tool name: EarthquakeTool, tool description: Provides real-time earthquake notifications and news.
\end{enumerate}
[List of Tools with Names and Descriptions End]
\vspace{2ex}

\textbf{Here are some examples:} [Examples Start] \\
query: ``I'm planning a hiking trip next week. What will the weather be like in the Grand Canyon?'' tool: WeatherTool \\
query: ``I'm a beginner surfer. Can you suggest a beach with mild waves for me to practice in Australia?'' tool: AusSurfReport \\
... \\
{[Examples End]}
\vspace{2ex}

\textbf{User query:} ``Planning a beach day next week, while considering the ideal weather conditions...'' \\
\textbf{tool:}
\end{tcolorbox}
\caption{MetaTool inference prompt with task instructions, candidate tools, and few-shot examples.}
\label{fig:metatool-eval-prompt}
\end{figure}

\begin{figure}[h]
\begin{tcolorbox}[title=API-Bank Inference Prompt, halign=left, boxrule=0.5pt]\footnotesize
\textbf{Instruction:} Based on the given API description and the existing conversation history 1..t, please generate the API request that the AI should call in step t+1 and output it in the format of [ApiName(key1='value1', key2='value2', ...)], replace the ApiName with the actual API name, and replace the key and value with the actual parameters.

\vspace{1ex}
Your output should start with a square bracket ``['' and end with a square bracket ``]''. Do not output any other explanation or prompt or the result of the API call in your output.

\vspace{1ex}
This year is 2023.

\vspace{2ex}
\textbf{API descriptions:} \\
\texttt{[\{"name": "GetWeather",} \\
\texttt{~~"description": "Get current weather information",} \\
\texttt{~~"parameters": \{"location": \{"type": "string",} \\
\texttt{~~~"description": "City name"\},} \\
\texttt{~~"unit": \{"type": "string",} \\
\texttt{~~~"enum": ["celsius", "fahrenheit"]\}\}\},} \\
\texttt{~...]}

\vspace{2ex}
\textbf{Input:} \\
User: What's the weather like in San Francisco? \\
AI: I'll check the current weather in San Francisco for you.

\vspace{2ex}
\textbf{Expected output:} \\
\texttt{[GetWeather(location='San Francisco', unit='fahrenheit')]}

\end{tcolorbox}
\caption{API-Bank inference prompt with API call generation instructions, schemas, and input dialogue.}
\label{fig:apibank-eval-prompt}
\end{figure}

\begin{figure}[h]
\begin{tcolorbox}[title=$\tau$-bench Inference Prompt, halign=left, boxrule=0.5pt]\footnotesize
\textbf{Instruction:} You need to act as an agent that uses the above tools to help the user according to the above policy. At each step, your generation should have exactly the following format:

\vspace{1ex}
\texttt{Thought: <A single line of reasoning to process the context} \\
\texttt{~~and inform the decision making. Do not include extra lines.>} \\
\texttt{Action: \{"name": <The name of the action>,} \\
\texttt{~~"arguments": <The arguments to the action in json format>\}}

\vspace{1ex}
The Action will be parsed, so it must be valid JSON. You should not use made-up or placeholder arguments.

\vspace{2ex}
\textbf{Available tools:} \\
\texttt{[\{"type": "function",} \\
\texttt{~~"function": \{"name": "get\_current\_weather",} \\
\texttt{~~~"description": "Get the current weather",} \\
\texttt{~~~"parameters": \{"type": "object",} \\
\texttt{~~~~"properties": \{"location": \{"type": "string",} \\
\texttt{~~~~~"description": "The city and state, e.g. San Francisco, CA"\},} \\
\texttt{~~~~"format": \{"type": "string", "enum": ["celsius", "fahrenheit"],} \\
\texttt{~~~~~"description": "The temperature unit to use. Infer this} \\
\texttt{~~~~~~from the users location."\}\},} \\
\texttt{~~~~"required": ["location", "format"]\}\}\},} \\
\texttt{~...]}

\vspace{2ex}
\textbf{User:} ``I want to know the current weather of San Francisco''

\vspace{2ex}
\textbf{Expected output:} \\
Thought: Since the user asks for the weather of San Francisco in USA, the unit should be in fahrenheit. I can query get\_current\_weather to get the weather. \\
Action: \{"name": "get\_current\_weather", "arguments": \{"location": "San Francisco, CA", "format": "fahrenheit"\}\}

\vspace{1ex}
\textit{[If tool returns "70F"]} \\
Thought: I can answer the user now. \\
Action: \{"name": "respond", "arguments": \{"response": "The current weather of San Francisco is 70F."\}\}

\end{tcolorbox}
\caption{$\tau$-Bench inference prompt with structured agent instructions, tool schemas, and multi-step reasoning examples.}
\label{fig:tau-bench-eval-prompt}
\end{figure}

\begin{figure}[h]
\centering
\begin{tcolorbox}[title=Error Classification Prompt, halign=left, boxrule=0.5pt]\footnotesize
You are an expert in analyzing tool selection errors in language models.
\vspace{2ex}

Given a query, the ground truth tool(s), the model's selected tool, and the available tool list, classify the error into ONE of these three categories:
\vspace{2ex}

\textbf{1. Schema Misalignment Error:} The model selected a tool name that seems plausible but doesn't exist in the tool list, OR the model selected a tool with a similar name pattern but different from the correct tool. This indicates the model is following its pretrained naming patterns rather than the actual schema.
\vspace{2ex}

\textbf{2. Functional Confusion Error:} The model selected an existing tool from the list that has similar functionality to the correct tool, but is not the right choice. The model understands the query but confuses functionally similar tools.
\vspace{2ex}

\textbf{3. Context Understanding Error:} The model failed to understand the query's intent or context, selecting a tool that is functionally unrelated to what the query requires.
\vspace{2ex}

Query: \texttt{\{query\}} \\
Ground Truth Tool: \texttt{\{ground\_truth\}} \\
Model Selected: \texttt{\{model\_output\}} \\
Available Tools: \texttt{\{tool\_list\}}
\vspace{2ex}

IMPORTANT: Respond with ONLY ONE of these exact phrases: \\
Schema Misalignment Error \\
Functional Confusion Error \\
Context Understanding Error
\end{tcolorbox}
\caption{Error classification prompt for analyzing tool selection errors.}
\label{fig:error-classification-prompt}
\end{figure}

\begin{figure}[h]
\centering
\begin{tcolorbox}[title=Training Data Generation Prompt, halign=left, boxrule=0.5pt]\footnotesize
You are an expert at explaining tool selection reasoning for AI systems. Your task is to generate high-quality, flowing reasoning that explains WHY specific tools were chosen for a given query.

\vspace{2ex}
\textbf{Your Task:}

\vspace{1ex}
Given:
\begin{itemize}
    \item A user's query
    \item A list of available tools with descriptions
    \item The tools that were selected (answer)
\end{itemize}

Generate: A clear, logical reasoning in natural paragraph form that explains why these specific tools are the best choices.

\vspace{2ex}
\textbf{Quality Criteria:}

\vspace{1ex}
Your reasoning MUST:
\begin{enumerate}
    \item Analyze the query deeply: Identify key requirements, implicit needs, and user intent
    \item Connect query to tools: Explicitly link query elements to tool capabilities
    \item Justify each selection: Explain why each tool is necessary
    \item Show tool synergy: Explain how the tools work together
    \item Consider alternatives: Briefly mention why other tools were NOT chosen
    \item Be specific: Use concrete details from the query and tool descriptions
    \item Write in flowing prose: Use natural paragraphs, not bullet points or lists
\end{enumerate}

\vspace{2ex}
\textbf{Example:}

\vspace{1ex}
...

\vspace{2ex}
\textbf{Now Generate:}

\vspace{1ex}
User's Query: \texttt{\{\{ query \}\}} \\
Selected Tools: \texttt{\{\{ selected\_tools \}\}}

\vspace{1ex}
Write your reasoning as flowing paragraphs that naturally cover query analysis, tool justification for each selected tool, their combined strategy, and why alternatives were not chosen.
\end{tcolorbox}
\caption{Training data generation prompt for supervised fine-tuning in Metatool.}
\label{fig:reasoning-generation-prompt}
\end{figure}

\begin{figure}[h]
\centering
\begin{tcolorbox}[title=Description Enhancement Prompt (EasyTool-style), halign=left, boxrule=0.5pt, colback=gray!5]\footnotesize
You are tasked with improving a tool description following the EasyTool methodology. The goal is to create clear, structured, and unified instructions that help language models better understand the tool's purpose and usage.
\vspace{1ex}

\textbf{Original Description:} \texttt{\{\{ description \}\}}
\vspace{1.5ex}

\textbf{Instructions:}
\begin{enumerate}[leftmargin=*, nosep, topsep=0pt]
    \item Create a CONCISE description that clearly states the tool's core functionality
    \item Focus on WHAT the tool does and WHEN to use it
    \item Remove redundant, verbose, or marketing-style language
    \item Use action-oriented verbs at the start (e.g., ``Retrieves'', ``Calculates'', ``Searches'')
    \item Keep the description to 1-2 sentences maximum
\end{enumerate}
\vspace{1ex}

\textbf{Critical Rules:}
\begin{itemize}[leftmargin=*, nosep, topsep=0pt]
    \item Do NOT include ANY tool name, API name, or brand name
    \item Do NOT start with ``This tool...'', ``The tool...'', or any name reference
    \item Start DIRECTLY with an action verb describing the functionality
\end{itemize}
\vspace{1.5ex}

\textbf{Example:} \\
\textit{Original:} ``WeatherAPI is a comprehensive weather tool that can be used to get current weather data and forecasts.'' \\
\textit{Improved:} ``Retrieves current weather conditions and forecasts including temperature, humidity, and precipitation for specified locations.''
\vspace{1ex}

Generate ONLY the improved description without additional explanation.

\end{tcolorbox}
\caption{Prompt template for description enhancement following EasyTool~\citep{yuan-etal-2025-easytool}.}
\label{fig:easytool-prompt}
\end{figure}

%=============================================

\end{document}